\documentclass{article}
\usepackage{iftex}
\ifPDFTeX
  \usepackage[T1]{fontenc}
  \usepackage[utf8]{inputenc}
\fi
\usepackage{longcat}
\usepackage[round,authoryear]{natbib}
\setcitestyle{authoryear,round,citesep={;},aysep={,},yysep={;}}
\usepackage{microtype}
\usepackage{graphicx}
\usepackage{amsmath,amssymb}
\usepackage{booktabs,multirow}
\usepackage{float}
\usepackage{enumitem}
\usepackage{subcaption}
\usepackage{xcolor}
\usepackage{colortbl}
\usepackage{hyperref}
\usepackage{url}
\hypersetup{
  colorlinks=true,
  linkcolor=blue,
  urlcolor=blue,
  citecolor=blue,
  pdftitle={Learning What to Practice: Diagnosis-Guided Self-Evolution for Language Models},
  pdfauthor={}
}

\title{Learning What to Practice: Diagnosis-Guided Self-Evolution for Language Models}
\author{%
  {\large
    \textbf{Xincheng Wei}\textsuperscript{1,2,*}\qquad
    \textbf{Yifan Ding}\textsuperscript{2,*}\qquad
    \textbf{Fucheng Xiong}\textsuperscript{3,*}
  }\\[0.65em]
  {\large
    \textbf{Yoshua Li}\textsuperscript{2,\textdagger}\quad
    \textbf{Dongsheng Ma}\textsuperscript{2,3}\quad
    \textbf{Rongxiang Weng}\textsuperscript{2}\quad
    \textbf{Xunliang Cai}\textsuperscript{2}
  }\\[0.45em]
  {\large
    \textbf{Wenjian Ding}\textsuperscript{4,\textdagger}\quad
    \textbf{Yao Zhang}\textsuperscript{5,\textdagger}
  }\\[1.0em]
  {\normalsize
    \textsuperscript{1}The Chinese University of Hong Kong, Shenzhen \\[0.25em]
    \textsuperscript{2}Meituan, LongCat Team \\[0.25em]
    \textsuperscript{3}Peking University \\[0.25em]
    \textsuperscript{4}Faculty of Health Data Science, Juntendo University, Chiba, Japan \\[0.25em]
    \textsuperscript{5}School of Statistics and Data Science, LPMC, KLMDASR \& AAIS, Nankai University, China
  }\\[0.75em]
  {\small
    \textsuperscript{*}Equal contribution.\qquad
    \textsuperscript{\textdagger}Corresponding authors: Wenjian Ding, Yao Zhang, Yoshua Li.
  }
}

\hypersetup{pdfauthor={Xincheng Wei, Yifan Ding, Fucheng Xiong, Yoshua Li, Dongsheng Ma, Rongxiang Weng, Xunliang Cai, Wenjian Ding, Yao Zhang}}

\renewcommand{\shorttitle}{DiagEvo}
\renewcommand{\headeright}{}

\begin{document}
\maketitle
\setcounter{footnote}{0}

\begin{abstract}
Self-play supports the self-evolution of language models, but solver performance can plateau or decline across rounds without guidance. Existing unguided methods typically use difficulty, learnability, or diversity signals to keep questions challenging and varied, without identifying which unresolved reasoning weaknesses to target. Existing guided methods rely on external task resources such as human examples, document corpora, or specified difficulty targets. We introduce \textbf{DiagEvo}, which guides question generation using the solver's failure history from self-play, without external task resources. Its diagnostician extracts recurring error causes and stores them in an error-cause memory. The memory groups related causes under skill nodes and tracks each as \textit{Active} or \textit{Mastered} according to self-consistency on targeted questions. The challenger uses these states and recurrence counts to balance cause-targeted generation with free exploration. \textit{Double-confidence filtering} retains intermediate-difficulty questions only when the most common solver answer has a clear vote lead. With the default 4B diagnostician, DiagEvo outperforms all baselines in mean accuracy across nine benchmarks for each solver: Qwen3-4B, Qwen3-8B, and OctoThinker-8B. On Qwen3-8B, DiagEvo reaches 72.3\% mean accuracy across five mathematical reasoning benchmarks, 4.5 percentage points above R-Zero. Its overall mean accuracy across nine benchmarks is 57.4\%, 3.5 percentage points above SPICE. Ablations show that mixed generation, memory-state updates with cross-state stitching, and double-confidence filtering contribute to these gains.
\end{abstract}

\suppressfloats[t]
\section{Introduction}

\begin{figure*}[t]
  \centering
  \includegraphics[width=\textwidth]{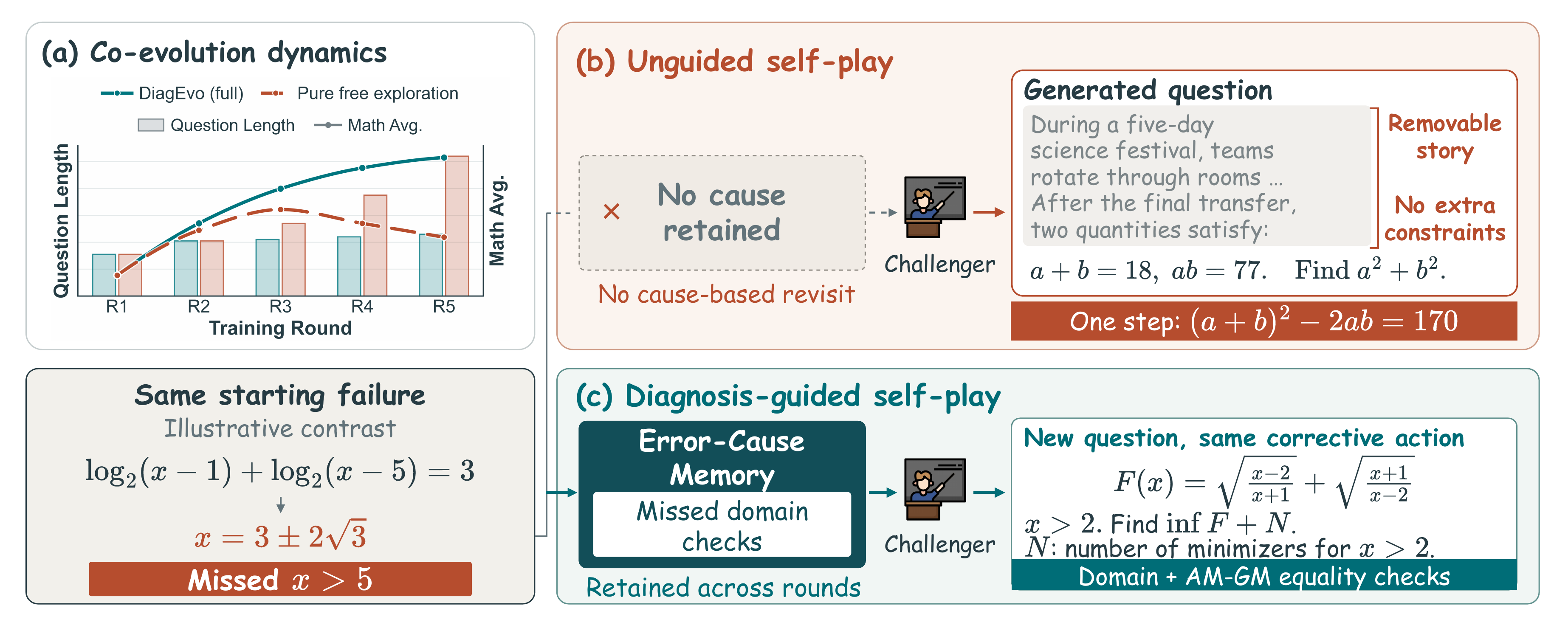}
  \caption{Co-evolution with and without error-cause memory. (a) Question length (bars) and mathematical average (lines). (b, c) An illustrative contrast from the same failure: unguided self-play adds story text without new constraints, while DiagEvo retains the error cause to guide questions requiring domain and AM-GM equality checks.}
  \label{fig:intro}
\end{figure*}

Recursive self-improvement (RSI) studies whether AI can improve how later systems are trained \citep{chi2026ai4aibench}. Recent work examines agents that design training algorithms or revise training data strategies \citep{chi2026ai4aibench,meng2026rsibenchdata}. Self-evolution is a related direction that aims to improve models with limited ongoing human supervision \citep{ref1}. Self-play provides a practical approach \citep{ref8,ref9}.

In self-play, the challenger generates questions and the solver learns to answer them. Alternating updates let questions evolve with the solver across rounds. Questions must stay near the solver's \textit{competence boundary}, exposing weaknesses while remaining learnable. Overly easy questions offer little improvement, while overly difficult ones produce unstable responses and weak training signals \citep{ref45}.

Existing self-play methods address this challenge in two ways. Label-free methods use solver responses to adjust difficulty \citep{ref8,ref9}. Different questions can expose the same error cause, so difficulty alone does not identify what the solver should practice next. Questions can grow longer without targeting these causes, and performance can stop improving (Figure~\ref{fig:intro}(a) and (b)) \citep{ref8,ref16}. Training on unreliable majority-vote pseudo-labels from difficult or unclear questions can also reinforce errors across rounds \citep{ref18}.

Guided self-play uses external task resources to stabilize training. R-Few uses human-annotated examples to guide question generation \citep{ref16}, while SPICE combines a document corpus with solver feedback to control question difficulty \citep{ref11}. DARC uses explicit difficulty targets, external documents, and a document-informed privileged teacher \citep{ref18}. Continued improvement thus depends on externally supplied task data or difficulty targets.

The solver's failure history offers a curriculum signal for future questions. Prior work on LLM agents stores past interactions in memory to guide later decisions or model updates \citep{ref38,ref39,ref40,ref41,ref42,ref43}. For the same question, we compare a failed trajectory that disagrees with the pseudo-label with one that agrees. Their reasoning differences reveal transferable error causes to target in new questions. The memory uses recurrence counts to prioritize causes. Self-consistency on targeted questions indicates when to reduce direct targeting as the solver evolves. The same error cause links diagnosis, question generation, and progress tracking across questions and rounds.

We propose \textbf{DiagEvo}, a diagnosis-guided self-play framework (Figure~\ref{fig:intro}(c)). A lightweight external LLM diagnostician extracts transferable error causes from solver trajectories produced during self-play and stores them in an error-cause memory. Each cause is \textit{Active} while directly targeted for generation and becomes \textit{Mastered} when the solver reaches high self-consistency on related questions. The challenger uses this memory to combine cause-targeted generation with free exploration. \textit{Double-confidence filtering} removes questions whose two leading answers have similar vote shares. The solver learns from the retained questions and pseudo-labels, and its failed training trajectories update the memory for the next round.

The main contributions are as follows.
\begin{enumerate}
\item We show that self-play can use its own training experience to guide what the solver practices next. Recurring failures reveal reasoning weaknesses that question difficulty alone does not identify.
\item We introduce DiagEvo, a diagnosis-guided self-play framework. Its error-cause memory connects failure diagnosis with question generation, adapting the curriculum to the solver's evolving weaknesses without external task resources.
\item With the default 4B diagnostician, DiagEvo achieves the best overall performance among the compared methods on Qwen3-4B, Qwen3-8B, and OctoThinker-8B. On Qwen3-8B, it reaches a 72.3\% mathematical reasoning average, 4.5 points above R-Zero, and exceeds DARC overall.
\end{enumerate}

\section{Method}
\label{sec:method}

\begin{figure*}[t]
  \centering
  \includegraphics[width=\textwidth]{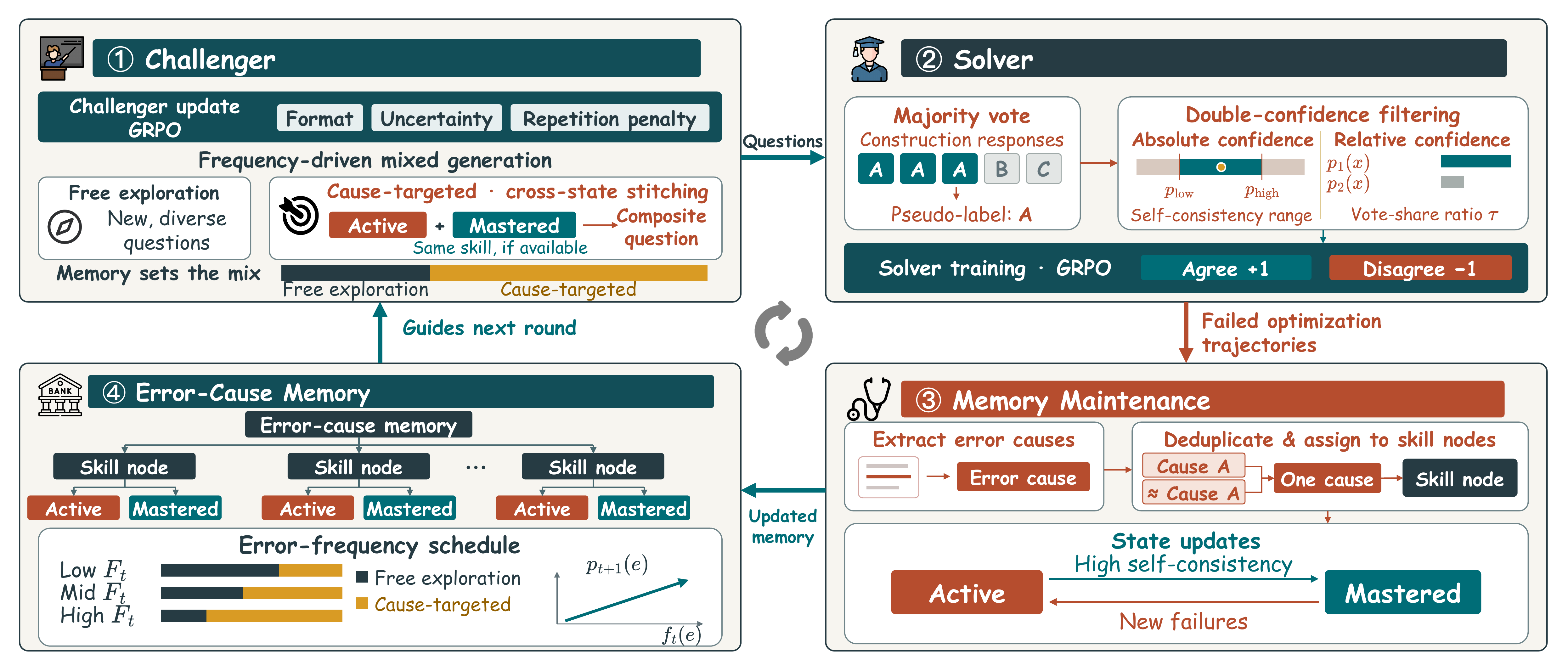}
  \caption{One DiagEvo round. (1) GRPO updates the challenger, which mixes free exploration with cause-targeted generation. (2) Double-confidence filtering selects question and pseudo-label pairs for solver training. (3) Memory maintenance compares failed optimization trajectories with construction references, then deduplicates, assigns, and updates the extracted causes. (4) The memory groups causes under skill nodes; their states and frequencies schedule the next round.}
  \label{fig:overview}
\end{figure*}

DiagEvo uses failed solver trajectories to guide question generation in the next round (Figure~\ref{fig:overview}). The challenger uses memory to generate curriculum candidates, and double-confidence filtering selects question and pseudo-label pairs for solver training. The solver's failed trajectories then update the memory.

The challenger and solver update in turn, with one frozen while the other updates. Only the challenger receives the memory.

\subsection{Diagnosis-Guided Question Generation}
\label{sec:generation}

Cause-targeted generation focuses on \textit{Active} causes in the previous round's memory. Free exploration maintains broad question coverage and can reveal unrecorded error causes.

\paragraph{Frequency-Driven Mixed Generation.}
Let $\mathcal{M}_t$ denote the error-cause memory committed after all updates in round $t$ (Section~\ref{sec:memory}). For a cause $e$, an active episode lasts from entry into \textit{Active} to promotion to \textit{Mastered}. Its frequency $f_t(e)$ counts diagnosed failures within the episode and resets to zero at promotion. The total $F_t=\sum_{e\in\mathcal{E}_t} f_t(e)$ sums frequencies across current active episodes.

Round 1 uses free exploration because the memory is empty. Its failure count $F_1$ is the fixed reference for $z_t=F_t/F_1$. The odds of cause-targeted generation relative to free exploration grow linearly with $z_t$, giving the next round's exploration probability and cause distribution:
\begin{equation}
  \frac{1-\varepsilon_{t+1}}{\varepsilon_{t+1}}
  = \frac{z_t}{k}
  = \frac{F_t}{kF_1},
  \;\Longrightarrow\;
  \varepsilon_{t+1}
  = \frac{1}{1+z_t/k}
  = \frac{F_1}{F_1+F_t/k},
  \qquad
  p_{t+1}(e) = \frac{f_t(e)}{F_t}.
  \label{eq:eps}
\end{equation}
The parameter $k$ balances the two modes, with equal probability at $z_t=k$. The distribution $p_{t+1}(e)$ samples \textit{Active} causes in proportion to their active-episode frequencies. If $F_t=0$, we use only free exploration, set $\varepsilon_{t+1}=1$, and do not compute $p_{t+1}$. More failures under \textit{Active} causes increase cause-targeted generation. Without reactivation by new failures, promotion reduces $F_t$ and shifts probability back to free exploration.

Let $q^{\mathrm{free}}_{\theta}(x)$ denote the challenger's free-exploration distribution and $q^{\mathrm{tar}}_{\theta}(x\mid e,\mathcal{M}_t)$ its distribution conditioned on an \textit{Active} cause $e$. For $F_t>0$, the complete question distribution is
\begin{equation}
  q_{t+1}(x\mid\mathcal{M}_t)
  = \varepsilon_{t+1}q^{\mathrm{free}}_{\theta}(x)
  + (1-\varepsilon_{t+1})\sum_{e\in\mathcal{E}_t}
  p_{t+1}(e)\,q^{\mathrm{tar}}_{\theta}(x\mid e,\mathcal{M}_t).
  \label{eq:mixed_generation}
\end{equation}
The mixture combines free exploration with targeting a cause sampled by frequency.

Within cause-targeted generation, cross-state stitching pairs the sampled \textit{Active} cause with a uniformly sampled \textit{Mastered} cause from the same skill node. The question requires skills from both causes. If no \textit{Mastered} sibling exists, the prompt uses only the \textit{Active} cause.

\paragraph{Challenger Update and Quality Reward.}
The current challenger $\theta_t$ samples an update batch $\mathcal{Q}^{\mathrm{chal}}_{t+1}$ from the mixed policy. Using the frozen solver's answers, GRPO updates the challenger to $\theta_{t+1}$ with the R-Zero reward \citep{ref8}. This reward favors questions near 50\% self-consistency and penalizes repetition. The updated challenger samples a fresh pool $\mathcal{Q}^{\mathrm{cur}}_{t+1}$ under the same memory schedule for filtering and solver training (Section~\ref{sec:solver}).

\subsection{Solver Optimization with Double-Confidence Filtering}
\label{sec:solver}

Training on $\mathcal{Q}^{\mathrm{cur}}_t$ requires reliable pseudo-labels. With ground-truth answers, intermediate success rates provide stronger RLVR learning signals \citep{ref45}. Generated questions lack these labels, so DiagEvo estimates difficulty from solver self-consistency. The absolute confidence constraint selects intermediate-difficulty questions but ignores the second most common answer. High-conflict questions have similar vote shares for their two leading answers. Their pseudo-labels can change across response samples. Training on them can reinforce errors across rounds. Double-confidence filtering compares these vote shares through a relative confidence constraint.

\paragraph{Dataset Construction and Double-Confidence Filtering.}
From this subsection onward, $\mathcal{Q}^{\mathrm{cur}}_t$ contains the candidates for solver round $t$. The solver samples $N$ construction responses per candidate $x$. Their majority answer gives the pseudo-label $\hat{y}(x)$. Their vote distribution determines filtering and provides evidence for state promotion (Section~\ref{sec:memory}).

Among construction responses, $p_1(x)$ and $p_2(x)$ are the vote shares of the majority and second most common answers, respectively. If no second answer exists, set $p_2(x)=0$. The self-consistency score $p(x)=p_1(x)$ measures answer agreement, not correctness. We retain questions satisfying both confidence constraints:
\begin{equation}
  \mathcal{D}_{\mathrm{train}}
  = \bigl\{\,x \bigm| p_{\mathrm{low}}\leq p(x)\leq p_{\mathrm{high}},
  \quad p_1(x)\geq\tau p_2(x)\,\bigr\}.
  \label{eq:filtering}
\end{equation}
The bounds $p_{\mathrm{low}}$ and $p_{\mathrm{high}}$ set the self-consistency range, while $\tau$ sets the required ratio of the two leading vote shares. For each retained question, we store one construction response that agrees with $\hat{y}(x)$ as the reference trajectory for diagnosis. GRPO \citep{ref28} samples $G$ separate optimization responses $\{r^{\mathrm{opt}}_j\}_{j=1}^{G}$. Each receives $+1$ for agreement with $\hat{y}(x)$ and $-1$ otherwise. These responses provide the policy-gradient signal. Their failed trajectories enter diagnosis (Section~\ref{sec:memory}).

\subsection{Error-Cause Memory and State Updates}
\label{sec:memory}

The memory groups related error causes under skill nodes. Cause states and frequencies guide question generation as the solver changes. We represent this two-level memory as
\begin{equation}
  \mathcal{M}_t = (\mathcal{V}_t,\mathcal{E}_t,a_t),
  \qquad v = (d_v,\mathbf{z}_v),
  \qquad e = (d_e,\mathbf{h}_e,s_t(e),f_t(e)).
  \label{eq:memory_representation}
\end{equation}
Here, $\mathcal{V}_t$ and $\mathcal{E}_t$ contain skill nodes and error causes, respectively. The mapping $a_t:\mathcal{E}_t\rightarrow\mathcal{V}_t$ assigns each cause to one skill node. Each node $v$ stores a description $d_v$ and an embedding $\mathbf{z}_v$. Causes store descriptions $d_e$, embeddings $\mathbf{h}_e$, curriculum states $s_t(e)\in\{\textit{Active},\textit{Mastered}\}$, and active-episode frequencies $f_t(e)$ (Section~\ref{sec:generation}).

Each round extracts and deduplicates causes, assigns new causes to skill nodes, and consolidates redundant nodes. The selected Qwen3-Instruct-2507 diagnostician makes all LLM-based semantic decisions using self-play data and the resulting memory. An embedding model supports retrieval within this memory. Both are frozen external models and receive no external task resources. Extraction runs independently for each failure. Later memory updates are sequential.

\paragraph{Comparative Error-Cause Extraction.}
After solver optimization in round $t$, diagnosis uses failed optimization trajectories from questions retained by Equation~\ref{eq:filtering}. For each question $x$, we pair a failed trajectory $r^{-}$ with its stored construction reference $r^{+}$. This reference remains available even when all optimization responses disagree with $\hat{y}(x)$. The diagnostician compares the pair to find their earliest reasoning difference:
\begin{equation}
  e^{\mathrm{new}}
  =D_{\mathrm{ext}}\bigl(x,\hat{y}(x),r^{-},r^{+}\bigr),
  \qquad r^{-}\not\sim\hat{y}(x),
  \quad r^{+}\sim\hat{y}(x).
  \label{eq:comparative_extraction}
\end{equation}
Here, $\sim$ denotes answer agreement. The extracted cause excludes question-specific values and surface form. This comparison identifies a relative reasoning difference without verifying ground-truth correctness.

The set $\mathcal{E}^{\mathrm{known}}(x)$ contains the causes supplied during generation: none for free exploration, the sampled \textit{Active} cause for single-cause generation, or both the \textit{Active} and stitched \textit{Mastered} causes for cross-state generation. The diagnostician compares the earliest reasoning difference with every supplied cause and assigns each failure to at most one match. A confirmed match adds one to the cause's failure count $\Delta_t(e)$ and reactivates it if \textit{Mastered}. If none matches, Equation~\ref{eq:comparative_extraction} produces a new candidate cause for memory matching.

\paragraph{Deduplication, Assignment, and Consolidation.}
Memory maintenance retrieves similar stored causes by embedding. The diagnostician confirms matches despite wording differences, assigns each observation to the matched cause, and adds one to $\Delta_t(e)$. It assigns unmatched candidates to a retrieved or new skill node. New causes start as \textit{Active} with frequency one. At round end, node embeddings retrieve potentially redundant skill nodes. The diagnostician confirms merges while preserving cause states and frequencies. Appendix~\ref{sec:appendix_prompts} provides prompt templates and retrieval and routing details.

\paragraph{Joint State and Frequency Update.}
An \textit{Active} cause is eligible for direct targeting. \textit{Mastered} is a curriculum state based on high self-consistency, not verified correctness. Causes in this state remain available for cross-state stitching.

For each cause present at the start of round $t$, we first apply promotion to its state from round $t-1$. The set $\mathcal{P}_t(e)\subseteq\mathcal{Q}^{\mathrm{cur}}_t$ contains all candidates targeting an \textit{Active} cause $e$, including those removed by filtering. Their mean self-consistency gives the promotion score:
\begin{equation}
  \mathrm{acc}_t(e)
  = \frac{1}{|\mathcal{P}_t(e)|}
    \sum_{x\in\mathcal{P}_t(e)} p_1(x),
  \qquad |\mathcal{P}_t(e)|>0.
  \label{eq:promotion_score}
\end{equation}
Using the promotion threshold $\theta_{\mathrm{up}}$, we record the provisional state and frequency as
\begin{equation}
  (\tilde{s}_t(e),\tilde{f}_t(e)) =
  \begin{cases}
    (\textit{Mastered},0),
      & s_{t-1}(e)=\textit{Active},\quad |\mathcal{P}_t(e)|>0,
        \quad \mathrm{acc}_t(e)\geq\theta_{\mathrm{up}}, \\[2pt]
    (s_{t-1}(e),f_{t-1}(e)),
      & \text{otherwise}.
  \end{cases}
  \label{eq:provisional_promotion}
\end{equation}
New optimization failures can reactivate a cause promoted earlier in the same round.

After solver optimization, $\Delta_t(e)$ counts failed optimization trajectories from retained questions assigned to $e$. Each failure is assigned to at most one cause through recurrence attribution or memory matching. For existing causes, the committed update is
\begin{equation}
  (s_t(e),f_t(e)) =
  \begin{cases}
    (\tilde{s}_t(e),\tilde{f}_t(e)),
      & \Delta_t(e)=0, \\[2pt]
    (\textit{Active},\tilde{f}_t(e)+\Delta_t(e)),
      & \Delta_t(e)>0,\quad\tilde{s}_t(e)=\textit{Active}, \\[2pt]
    (\textit{Active},\Delta_t(e)),
      & \Delta_t(e)>0,\quad\tilde{s}_t(e)=\textit{Mastered}.
  \end{cases}
  \label{eq:state_frequency_transition}
\end{equation}
\textit{Active} causes accumulate failures within their current episode. Reactivated causes count only new failures, excluding those from closed episodes. The committed states and frequencies determine $F_t$, $\varepsilon_{t+1}$, and $p_{t+1}$ for the next round.

\section{Experiments and Analysis}

We conduct experiments to answer three questions. Section~\ref{sec:setup} describes the models, baselines, benchmarks, and evaluation protocol. How does DiagEvo compare with prior methods across solvers, diagnostician scales, and reasoning domains (Section~\ref{sec:main_results})? How do mixed generation, memory-state updates with cross-state stitching, and double-confidence filtering affect performance (Section~\ref{sec:ablation})? How do solver performance, error-cause memory, and pseudo-label quality change across rounds (Section~\ref{sec:dynamics})? A case study traces how a diagnosed failure shapes later questions and solver behavior (Section~\ref{sec:case_study}).

\subsection{Experimental Setup}
\label{sec:setup}

\paragraph{Models.}
We evaluate DiagEvo on Qwen3-4B-Base, Qwen3-8B-Base, and OctoThinker-8B-Hybrid-Base. For each solver, we test Qwen3-4B-Instruct-2507, Qwen3-30B-A3B-Instruct-2507, and Qwen3-235B-A22B-Instruct-2507 as diagnosticians.

\paragraph{Baselines.}
We compare DiagEvo with base models, R-Zero \citep{ref8}, Absolute Zero \citep{ref9}, SPICE \citep{ref11}, R-Few \citep{ref16}, and DARC \citep{ref18}. R-Zero, Absolute Zero, and DiagEvo are label-free: they construct curricula and update model parameters without external task resources, while external models are allowed. We evaluate the base models and reproduce R-Zero. Other baseline scores come from DARC \citep{ref18} under the same evaluation protocol.

\paragraph{Evaluation Data and Protocol.}
We evaluate mathematical reasoning on MATH-500 \citep{ref29}, GSM8K \citep{ref30}, OlympiadBench \citep{ref31}, Minerva Math \citep{ref32}, and AMC \citep{ref33}. General reasoning benchmarks are MMLU-Pro \citep{ref34}, SuperGPQA \citep{ref35}, GPQA-Diamond \citep{ref36}, and BBEH \citep{ref37}. We report accuracy (mean@32 for AMC). We tune hyperparameters on a fixed, held-out validation set of 600 labeled questions, split equally between MATH training and MMLU validation. Each DiagEvo run uses the checkpoint with the highest validation accuracy for all nine test benchmarks. We report means and standard deviations over three independent runs; Appendix~\ref{sec:appendix_repro} details evaluation and baseline score validation.

\subsection{Main Results}
\label{sec:main_results}

\begin{table*}[t]
  \centering
  \caption{Reasoning accuracy (\%). DiagEvo reports mean $\pm$ standard deviation over three runs; aggregate columns show means only. `Diag.' denotes the Qwen3-Instruct-2507 diagnostician. Blue shades mark the top three means per column and block, darkest first. Bold and underlining mark first and second, including ties. $\dagger$ denotes results reported in DARC \citep{ref18}; others are ours.}
  \label{tab:main_results}
  \definecolor{TableRule}{HTML}{687786}
  \definecolor{TableMuted}{HTML}{334155}
  \definecolor{TableSolver}{HTML}{EDF0F4}
  \definecolor{RankFirst}{HTML}{73B3D8}
  \definecolor{RankSecond}{HTML}{B1D3E8}
  \definecolor{RankThird}{HTML}{E4F0F8}
  \newcommand{\scorecell}[2]{%
    \mbox{#1\kern0.6pt{\fontsize{5.8}{6.4}\selectfont\color{TableMuted}$\pm$\kern0.2pt#2}}}
  \newcommand{\diaglabel}[1]{%
    \mbox{\fontsize{7.6}{8.4}\selectfont DiagEvo {\bfseries(Diag.\ #1)}}}
  \newcommand{\benchhead}[1]{%
    \mbox{\fontsize{7.6}{8.6}\selectfont#1}}
  \fontsize{8.2}{9.4}\selectfont
  \setlength{\tabcolsep}{1.2pt}
  \renewcommand{\arraystretch}{1.05}
  \arrayrulecolor{TableRule}
  \resizebox{\textwidth}{!}{%
  \begin{tabular}{l ccccc c @{\hspace{4pt}} cccc c @{\hspace{4pt}} c}
    \toprule[0.8pt]
    \multirow{2}{*}{\shortstack[l]{\textbf{Method}\\{\fontsize{7}{8}\selectfont(Diagnostician)}}}
      & \multicolumn{6}{c}{\textbf{Mathematical Reasoning}}
      & \multicolumn{5}{c}{\textbf{General Reasoning}}
      & \multirow{2}{*}{\textbf{Overall}} \\
    \cmidrule(lr){2-7} \cmidrule(lr){8-12}
    & \benchhead{AMC} & \benchhead{Minerva} & \benchhead{MATH-500}
      & \benchhead{GSM8K} & \benchhead{Olympiad} & \benchhead{\textbf{Avg.}}
      & \benchhead{MMLU-Pro} & \benchhead{SuperGPQA} & \benchhead{GPQA-D}
      & \benchhead{BBEH} & \benchhead{\textbf{Avg.}} & \\
    \midrule
    \rowcolor{TableSolver}
    \multicolumn{13}{l}{\rule{0pt}{10.8pt}\textbf{Qwen3-4B-Base}} \\[1pt]
    Base & 47.7 & 41.5 & 68.4 & 72.4 & 35.0 & 53.0 & 51.4 & 25.4 & 26.3 & 8.2 & 27.8 & 41.8 \\
    R-Zero & 47.8 & 50.4 & 74.4 & 90.1 & 40.3 & 60.6 & 53.9 & 27.7 & 35.9 & 10.2 & 31.9 & 47.9 \\
    Absolute Zero$^{\dagger}$ & 50.0 & 41.9 & 76.2 & 89.3 & 41.5 & 59.8 & 52.6 & 27.1 & 35.3 & 8.3 & 30.8 & 46.9 \\
    SPICE$^{\dagger}$ & 50.9 & 55.5 & 77.9 & 91.9 & 41.9 & 63.6 & \cellcolor{RankThird}56.5 & 28.3 & 37.9 & \cellcolor{RankSecond}\underline{11.3} & 33.5 & 50.2 \\
    R-Few (1\%)$^{\dagger}$ & 52.7 & 52.1 & 77.8 & \cellcolor{RankSecond}\underline{92.3} & 42.4 & 63.5 & 55.9 & 29.4 & 35.4 & \cellcolor{RankThird}11.2 & 33.0 & 49.9 \\
    DARC$^{\dagger}$ & 60.3 & 57.7 & 77.6 & 91.9 & 45.8 & 66.7 & \cellcolor{RankSecond}\underline{56.9} & 29.2 & 38.9 & \cellcolor{RankThird}11.2 & 34.1 & 52.2 \\
    \addlinespace[1.5pt]
    \diaglabel{4B} & \cellcolor{RankThird}\scorecell{62.2}{0.3} & \cellcolor{RankThird}\scorecell{60.7}{1.0} & \cellcolor{RankThird}\scorecell{78.2}{0.2} & \cellcolor{RankThird}\scorecell{92.0}{0.3} & \cellcolor{RankThird}\scorecell{48.3}{0.5} & \cellcolor{RankThird}68.3 & \cellcolor{RankFirst}\scorecell{\textbf{57.0}}{0.4} & \cellcolor{RankThird}\scorecell{30.3}{0.4} & \cellcolor{RankThird}\scorecell{41.2}{1.2} & \cellcolor{RankFirst}\scorecell{\textbf{11.5}}{0.4} & \cellcolor{RankThird}35.0 & \cellcolor{RankThird}53.5 \\[1pt]
    \diaglabel{30B-A3B} & \cellcolor{RankSecond}\scorecell{\underline{62.5}}{0.5} & \cellcolor{RankSecond}\scorecell{\underline{61.4}}{0.6} & \cellcolor{RankSecond}\scorecell{\underline{78.5}}{0.6} & \scorecell{91.9}{0.2} & \cellcolor{RankSecond}\scorecell{\underline{50.1}}{0.7} & \cellcolor{RankSecond}\underline{68.9} & \cellcolor{RankSecond}\scorecell{\underline{56.9}}{0.1} & \cellcolor{RankFirst}\scorecell{\textbf{30.7}}{0.5} & \cellcolor{RankFirst}\scorecell{\textbf{42.1}}{0.8} & \cellcolor{RankSecond}\scorecell{\underline{11.3}}{0.3} & \cellcolor{RankFirst}\textbf{35.3} & \cellcolor{RankSecond}\underline{53.9} \\[1pt]
    \diaglabel{235B-A22B} & \cellcolor{RankFirst}\scorecell{\textbf{63.0}}{0.4} & \cellcolor{RankFirst}\scorecell{\textbf{61.6}}{0.8} & \cellcolor{RankFirst}\scorecell{\textbf{78.7}}{0.4} & \cellcolor{RankFirst}\scorecell{\textbf{92.4}}{0.3} & \cellcolor{RankFirst}\scorecell{\textbf{50.7}}{0.4} & \cellcolor{RankFirst}\textbf{69.3} & \cellcolor{RankFirst}\scorecell{\textbf{57.0}}{0.6} & \cellcolor{RankSecond}\scorecell{\underline{30.6}}{0.4} & \cellcolor{RankSecond}\scorecell{\underline{41.9}}{0.9} & \cellcolor{RankSecond}\scorecell{\underline{11.3}}{0.3} & \cellcolor{RankSecond}\underline{35.2} & \cellcolor{RankFirst}\textbf{54.1} \\[1pt]
    \addlinespace[3pt]
    \midrule
    \rowcolor{TableSolver}
    \multicolumn{13}{l}{\rule{0pt}{10.8pt}\textbf{Qwen3-8B-Base}} \\[1pt]
    Base & 61.7 & 50.0 & 74.2 & 91.1 & 40.1 & 63.4 & 58.1 & 30.2 & 33.8 & 10.5 & 33.2 & 50.0 \\
    R-Zero & 62.7 & 59.6 & 80.8 & 92.3 & 43.6 & 67.8 & 61.7 & 31.8 & 39.9 & 11.4 & 36.2 & 53.8 \\
    Absolute Zero$^{\dagger}$ & 62.5 & 52.9 & 76.6 & 92.0 & 47.8 & 66.4 & 62.5 & 33.5 & 36.8 & 10.8 & 35.9 & 52.8 \\
    SPICE$^{\dagger}$ & 60.9 & 55.2 & 81.4 & \cellcolor{RankThird}93.8 & 48.0 & 67.9 & 61.0 & 32.4 & 40.4 & 12.1 & 36.5 & 53.9 \\
    R-Few (1\%)$^{\dagger}$ & 69.3 & 59.6 & 81.6 & \cellcolor{RankSecond}\underline{94.0} & 44.0 & 69.7 & \cellcolor{RankThird}62.8 & 32.7 & 40.4 & 11.8 & 36.9 & 55.1 \\
    DARC$^{\dagger}$ & 68.9 & 61.4 & 83.0 & \cellcolor{RankSecond}\underline{94.0} & 48.4 & 71.1 & 62.3 & 32.8 & 44.4 & 11.8 & 37.8 & 56.3 \\
    \addlinespace[1.5pt]
    \diaglabel{4B} & \cellcolor{RankThird}\scorecell{69.6}{0.3} & \cellcolor{RankThird}\scorecell{63.2}{0.6} & \cellcolor{RankThird}\scorecell{83.3}{0.3} & \cellcolor{RankThird}\scorecell{93.8}{0.2} & \cellcolor{RankThird}\scorecell{51.4}{0.7} & \cellcolor{RankThird}72.3 & \cellcolor{RankThird}\scorecell{62.8}{0.7} & \cellcolor{RankThird}\scorecell{33.8}{0.6} & \cellcolor{RankThird}\scorecell{46.0}{1.3} & \cellcolor{RankThird}\scorecell{12.4}{0.5} & \cellcolor{RankThird}38.8 & \cellcolor{RankThird}57.4 \\[1pt]
    \diaglabel{30B-A3B} & \cellcolor{RankSecond}\scorecell{\underline{69.8}}{0.2} & \cellcolor{RankSecond}\scorecell{\underline{64.1}}{0.4} & \cellcolor{RankSecond}\scorecell{\underline{83.5}}{0.7} & \cellcolor{RankThird}\scorecell{93.8}{0.5} & \cellcolor{RankSecond}\scorecell{\underline{53.5}}{0.4} & \cellcolor{RankSecond}\underline{72.9} & \cellcolor{RankSecond}\scorecell{\underline{63.0}}{0.2} & \cellcolor{RankSecond}\scorecell{\underline{33.9}}{0.3} & \cellcolor{RankSecond}\scorecell{\underline{47.3}}{0.6} & \cellcolor{RankFirst}\scorecell{\textbf{12.6}}{0.1} & \cellcolor{RankSecond}\underline{39.2} & \cellcolor{RankSecond}\underline{57.9} \\[1pt]
    \diaglabel{235B-A22B} & \cellcolor{RankFirst}\scorecell{\textbf{70.0}}{0.5} & \cellcolor{RankFirst}\scorecell{\textbf{64.8}}{0.9} & \cellcolor{RankFirst}\scorecell{\textbf{83.8}}{0.9} & \cellcolor{RankFirst}\scorecell{\textbf{94.1}}{0.4} & \cellcolor{RankFirst}\scorecell{\textbf{53.8}}{0.6} & \cellcolor{RankFirst}\textbf{73.3} & \cellcolor{RankFirst}\scorecell{\textbf{63.1}}{0.4} & \cellcolor{RankFirst}\scorecell{\textbf{34.0}}{0.4} & \cellcolor{RankFirst}\scorecell{\textbf{47.5}}{0.9} & \cellcolor{RankSecond}\scorecell{\underline{12.5}}{0.3} & \cellcolor{RankFirst}\textbf{39.3} & \cellcolor{RankFirst}\textbf{58.2} \\[1pt]
    \addlinespace[3pt]
    \midrule
    \rowcolor{TableSolver}
    \multicolumn{13}{l}{\rule{0pt}{10.8pt}\textbf{OctoThinker-8B-Hybrid-Base}} \\[1pt]
    Base & 27.7 & 21.7 & 44.6 & 68.4 & 16.9 & 35.9 & 14.9 & 11.3 & 15.7 & 0.7 & 10.7 & 24.7 \\
    R-Zero & 32.2 & 32.7 & 58.0 & 84.7 & 22.4 & 46.0 & 37.1 & 17.7 & 21.2 & 7.7 & 20.9 & 34.9 \\
    Absolute Zero$^{\dagger}$ & 32.5 & 34.9 & 56.8 & 87.0 & 25.6 & 47.4 & 31.4 & 18.8 & 27.8 & 5.0 & 20.8 & 35.5 \\
    SPICE$^{\dagger}$ & \cellcolor{RankFirst}\textbf{35.2} & 40.8 & 58.4 & 87.3 & 25.6 & 49.5 & 41.3 & 19.9 & 29.8 & 7.2 & 24.6 & 38.4 \\
    DARC$^{\dagger}$ & 31.9 & 43.0 & 62.4 & \cellcolor{RankThird}88.0 & 30.7 & 51.2 & 43.8 & 22.3 & 32.3 & \cellcolor{RankThird}10.8 & \cellcolor{RankThird}27.3 & 40.6 \\
    \addlinespace[1.5pt]
    \diaglabel{4B} & \cellcolor{RankThird}\scorecell{33.7}{0.7} & \cellcolor{RankThird}\scorecell{45.2}{1.1} & \cellcolor{RankThird}\scorecell{62.8}{0.6} & \cellcolor{RankSecond}\scorecell{\underline{88.2}}{0.2} & \cellcolor{RankThird}\scorecell{33.6}{0.8} & \cellcolor{RankThird}52.7 & \cellcolor{RankThird}\scorecell{44.3}{0.5} & \cellcolor{RankThird}\scorecell{23.1}{0.4} & \cellcolor{RankThird}\scorecell{35.0}{1.1} & \cellcolor{RankSecond}\scorecell{\underline{11.2}}{0.2} & \cellcolor{RankSecond}\underline{28.4} & \cellcolor{RankThird}41.9 \\[1pt]
    \diaglabel{30B-A3B} & \cellcolor{RankThird}\scorecell{33.7}{0.4} & \cellcolor{RankSecond}\scorecell{\underline{46.4}}{0.6} & \cellcolor{RankSecond}\scorecell{\underline{63.4}}{0.3} & \cellcolor{RankSecond}\scorecell{\underline{88.2}}{0.4} & \cellcolor{RankSecond}\scorecell{\underline{36.0}}{0.7} & \cellcolor{RankSecond}\underline{53.5} & \cellcolor{RankFirst}\scorecell{\textbf{44.8}}{0.2} & \cellcolor{RankSecond}\scorecell{\underline{23.5}}{0.2} & \cellcolor{RankSecond}\scorecell{\underline{35.4}}{0.5} & \cellcolor{RankFirst}\scorecell{\textbf{11.3}}{0.2} & \cellcolor{RankFirst}\textbf{28.8} & \cellcolor{RankSecond}\underline{42.5} \\[1pt]
    \diaglabel{235B-A22B} & \cellcolor{RankSecond}\scorecell{\underline{33.9}}{0.4} & \cellcolor{RankFirst}\scorecell{\textbf{47.2}}{0.8} & \cellcolor{RankFirst}\scorecell{\textbf{63.7}}{0.4} & \cellcolor{RankFirst}\scorecell{\textbf{88.5}}{0.5} & \cellcolor{RankFirst}\scorecell{\textbf{36.4}}{0.4} & \cellcolor{RankFirst}\textbf{53.9} & \cellcolor{RankSecond}\scorecell{\underline{44.6}}{0.3} & \cellcolor{RankFirst}\scorecell{\textbf{23.6}}{0.3} & \cellcolor{RankFirst}\scorecell{\textbf{35.5}}{0.8} & \cellcolor{RankFirst}\scorecell{\textbf{11.3}}{0.4} & \cellcolor{RankFirst}\textbf{28.8} & \cellcolor{RankFirst}\textbf{42.7} \\[1pt]
    \bottomrule[0.8pt]
  \end{tabular}%
  }
  \arrayrulecolor{black}
\end{table*}

\paragraph{Comparison with Prior Methods.}
With the default 4B diagnostician, DiagEvo ranks first overall on all three solvers, exceeding every baseline by at least 1.1 points. On Qwen3-8B, its mathematical and general averages reach 72.3\% and 38.8\%. These exceed R-Zero by 4.5 and 2.6 points, and DARC by 1.2 and 1.0 points. DiagEvo achieves these gains using self-generated training questions and pseudo-labels.

Overall scores are 53.5\% on Qwen3-4B and 41.9\% on OctoThinker-8B, both 1.3 points above DARC.

\paragraph{Effect of Diagnostician Scale.}
Larger diagnosticians produce consistent but modest gains in mathematical reasoning. Scaling from 4B to 235B-A22B raises the mathematical average by 1.0 point on both Qwen3 solvers and 1.2 points on OctoThinker-8B. The largest gains appear on harder mathematical benchmarks. On OlympiadBench, the gains are 2.4, 2.4, and 2.8 points across the three solvers.

General averages change by only 0.2 to 0.5 points, with no gain on some benchmarks. Diagnostician scale is therefore not the main source of DiagEvo's gains.

\paragraph{Cross-Domain Generalization.}
The solver trains only on mathematical questions, so gains on general-reasoning benchmarks measure cross-domain transfer. With the default diagnostician, general averages rise from 33.2\% to 38.8\% on Qwen3-8B and from 27.8\% to 35.0\% on Qwen3-4B. OctoThinker-8B improves from 10.7\% to 28.4\%. These gains span different solver scales and architectures.

\paragraph{Training Time Breakdown.}
With the default 4B diagnostician, diagnosis and memory maintenance account for 5.7\% of total training time. Table~\ref{tab:appendix_runtime} reports the breakdown by component.

\subsection{Ablation Studies}
\label{sec:ablation}

We evaluate the mixed generation policy, memory-state updates with cross-state stitching, and double-confidence filtering on Qwen3-8B-Base (Table~\ref{tab:ablation}). Component ablations keep the remaining hyperparameters at the full method's defaults. Each run uses the checkpoint with the highest validation accuracy. DiagEvo (full) and pure free exploration are each trained for seven rounds.

\paragraph{Question Generation Policy.}

\begin{table}[!t]
  \centering
  \caption{Ablations on Qwen3-8B-Base. DiagEvo (full) is the shared reference, using $\tau=1.6$ for double-confidence filtering. Each variant changes one component. Subscripts show changes from full in percentage points.}
  \label{tab:ablation}
  \definecolor{AblationDelta}{HTML}{A34D45}
  \definecolor{AblationRule}{HTML}{687786}
  \newcommand{\ablationscore}[2]{%
    \mbox{#1\kern0.6pt\raisebox{-1.1pt}{\fontsize{5.6}{6.2}\selectfont\color{AblationDelta}$-$#2}}}
  \newcommand{\ablationgroup}[1]{{\fontsize{7.6}{8.4}\selectfont\bfseries\shortstack{#1}}}
  \newcommand{\ablationhead}[1]{{\fontsize{7.4}{8.2}\selectfont\shortstack{#1}}}
  \fontsize{8.2}{9.4}\selectfont
  \setlength{\tabcolsep}{1.6pt}
  \renewcommand{\arraystretch}{1.05}
  \arrayrulecolor{AblationRule}
  \resizebox{\textwidth}{!}{%
  \begin{tabular}{l ccc @{\hspace{4pt}} ccc @{\hspace{4pt}} cc @{\hspace{4pt}} c}
    \toprule
    & \multicolumn{3}{c}{\ablationgroup{Question generation\\policy}}
      & \multicolumn{3}{c}{\ablationgroup{Memory-state updates\\and cross-state stitching}}
      & \multicolumn{2}{c}{\ablationgroup{Pseudo-label\\filtering}}
      & \ablationgroup{Reference} \\
    \cmidrule(lr){2-4} \cmidrule(lr){5-7} \cmidrule(lr){8-9} \cmidrule(lr){10-10}
    \textbf{Metric}
      & \ablationhead{Frozen\\challenger}
      & \ablationhead{Pure free\\exploration}
      & \ablationhead{Pure cause-targeted\\generation}
      & \ablationhead{No state\\updates}
      & \ablationhead{No cross-state\\stitching}
      & \ablationhead{Random-pair\\stitching}
      & \ablationhead{No\\filtering}
      & \ablationhead{Absolute confidence\\constraint only}
      & \ablationhead{\textbf{DiagEvo}\\\textbf{(full)}} \\
    \midrule
    Math Avg. & \ablationscore{68.5}{3.8} & \ablationscore{69.5}{2.8} & \ablationscore{70.1}{2.2} & \ablationscore{70.1}{2.2} & \ablationscore{70.8}{1.5} & \ablationscore{71.3}{1.0} & \ablationscore{69.4}{2.9} & \ablationscore{70.9}{1.4} & \textbf{72.3} \\
    General Avg. & \ablationscore{36.9}{1.9} & \ablationscore{37.2}{1.6} & \ablationscore{37.6}{1.2} & \ablationscore{37.1}{1.7} & \ablationscore{37.5}{1.3} & \ablationscore{37.9}{0.9} & \ablationscore{36.7}{2.1} & \ablationscore{37.5}{1.3} & \textbf{38.8} \\
    \bottomrule
  \end{tabular}%
  }
  \arrayrulecolor{black}
\end{table}

Freezing the challenger reduces the mathematical average by 3.8 points, the largest drop in this group. The full policy outperforms both single-mode variants, showing that free exploration and cause-targeted generation are complementary. Its frequency-driven schedule exceeds the best tested fixed ratio by 1.1 points in mathematical average and 1.2 in general average (Table~\ref{tab:ablation_fixed_ratio}).

\paragraph{Memory-State Updates and Cross-State Stitching.}

Without state updates, all causes remain \textit{Active}, so cross-state stitching is unavailable. Enabling promotion and reactivation without stitching raises mathematical and general averages by 0.7 and 0.4 points, respectively. Adding cross-state stitching yields a further 1.5 and 1.3 points. Selecting the \textit{Mastered} cause from the same skill node also improves the mathematical average by 1.0 point over random-pair stitching.

\paragraph{Pseudo-Label Filtering.}

The absolute confidence constraint improves both averages over no filtering. Adding the relative confidence constraint raises the mathematical average by a further 1.4 points, showing that both constraints contribute. The mathematical average varies by at most 1.0 point across the tested values of either $k$ or $\tau$ (Appendix~\ref{sec:appendix_sensitivity}).

\subsection{Co-Evolution Dynamics}
\label{sec:dynamics}

\begin{figure*}[!t]
  \centering
  \begin{subfigure}{0.328\textwidth}
    \centering
    \includegraphics[width=\linewidth]{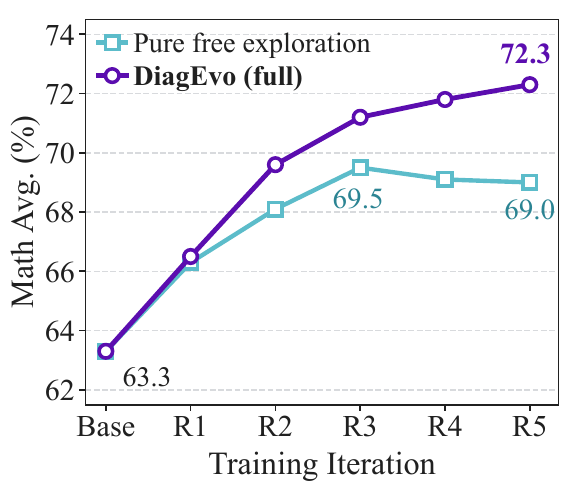}
    \caption{Per-round performance}
    \label{fig:dynamics_perf}
  \end{subfigure}\hfill
  \begin{subfigure}{0.328\textwidth}
    \centering
    \includegraphics[width=\linewidth]{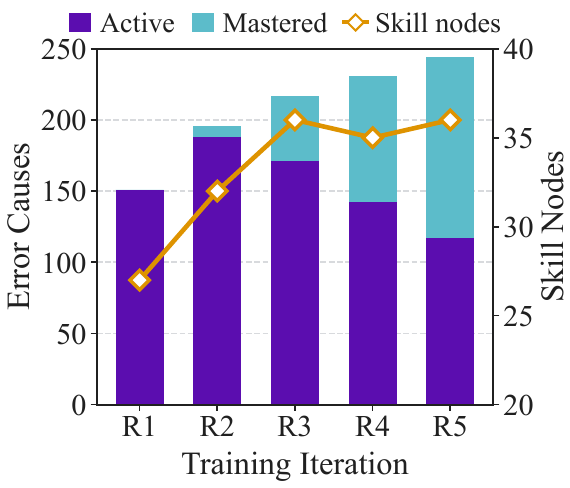}
    \caption{Memory lifecycle}
    \label{fig:dynamics_memory}
  \end{subfigure}\hfill
  \begin{subfigure}{0.328\textwidth}
    \centering
    \includegraphics[width=\linewidth]{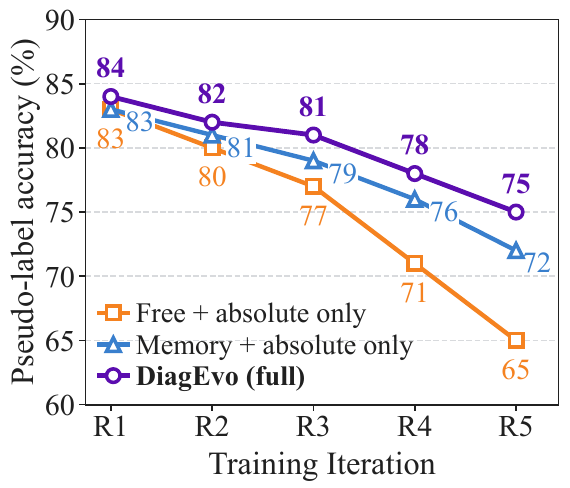}
    \caption{Pseudo-label accuracy}
    \label{fig:dynamics_label}
  \end{subfigure}

  \caption{Co-evolution on Qwen3-8B-Base. (a) Mathematical averages, both with double-confidence filtering. (b) Counts of \textit{Active} and \textit{Mastered} causes and skill nodes. (c) Pseudo-label accuracy (\%) after each round's training on 200 shared questions. Starting with pure free exploration and only the absolute confidence constraint, the three settings add memory, then the relative confidence constraint.}
  \label{fig:dynamics}
\end{figure*}

\paragraph{Solver Performance and Memory Dynamics.}

DiagEvo's mathematical average improves in each of the first five rounds, reaching 72.3\% (Figure~\ref{fig:dynamics}(a)). It then falls to 72.0\% and 71.4\% in rounds 6 and 7. All three DiagEvo runs in this setting also attain their highest validation accuracy at round 5. Pure free exploration peaks at 69.5\% in round 3, matching its ablation score (Table~\ref{tab:ablation}), then falls to 69.0\%, 68.6\%, and 68.0\% in rounds 5, 6, and 7.

New error causes enter each round, raising the total from 151 to 244 (Figure~\ref{fig:dynamics}(b)). End-of-round consolidation merges redundant skill nodes, whose count stabilizes near 36 after round 3. \textit{Active} causes peak at 188 in round 2 and fall to 117 by round 5. By then, \textit{Mastered} causes reach 127, over half of the memory. Promotion resets the closed episode's frequency and removes its past failures from $F_t$. The schedule then uses the updated $F_t$ to balance cause-targeted generation with free exploration (Equation~\ref{eq:eps}).

\paragraph{Pseudo-Label Quality.}

Figure~\ref{fig:dynamics}(c) compares the three solvers after each round's training on 200 shared challenger-generated questions. Each solver forms a pseudo-label by majority vote over 12 responses per question, matching the response count for curriculum construction. Qwen3.8-Max proposes reference answers, which humans verify and correct. Pseudo-label accuracy is the percentage of pseudo-labels matching these human-verified oracle answers.

With pure free exploration and only the absolute confidence constraint, pseudo-label accuracy falls from 83\% in round 1 to 65\% in round 5. On the same round-5 questions, training with memory under the same constraint raises accuracy to 72\%. Adding the relative confidence constraint during training raises it to 75\%. These gains of 7 and 3 points support combining memory and double-confidence filtering during training to improve pseudo-label accuracy on shared questions. Appendix~\ref{sec:appendix_analysis} details the evaluation protocols and analyzes question coverage, length, lexical diversity, and difficulty across rounds.

\subsection{Case Study}
\label{sec:case_study}

\begin{figure*}[!t]
  \centering
  \includegraphics[width=\textwidth]{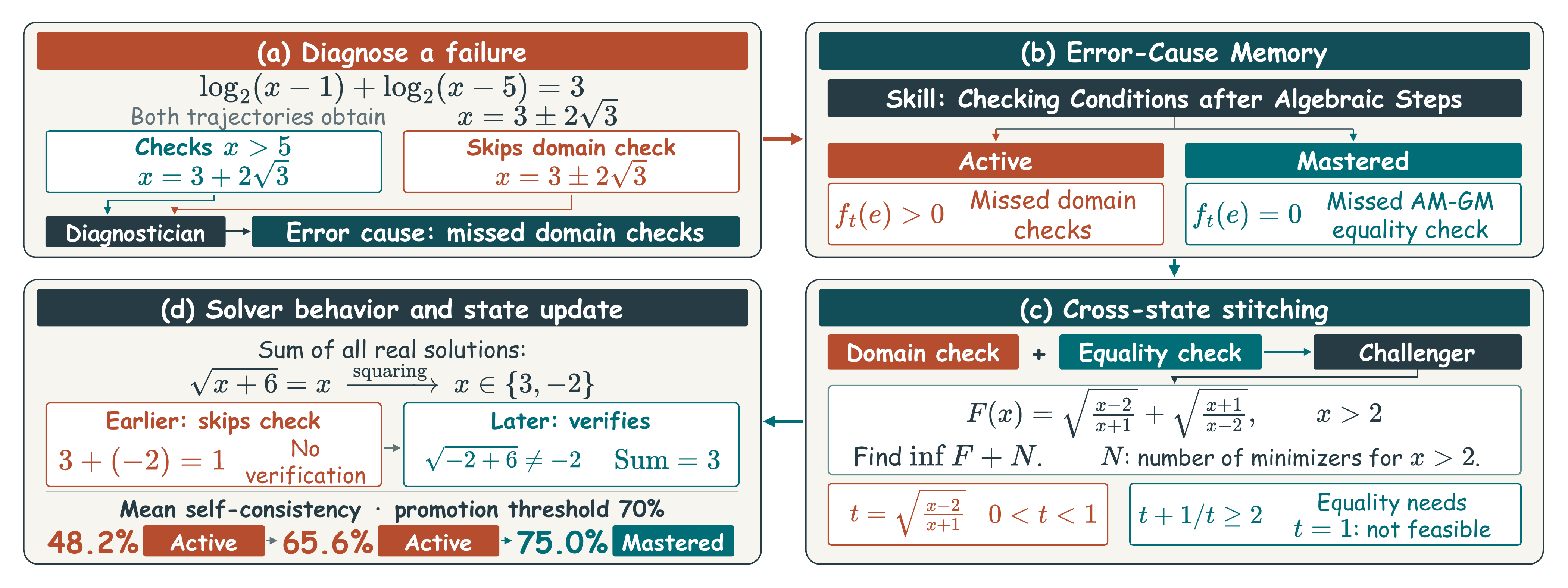}
  \caption{Case study on Qwen3-8B-Base. (a, b) Diagnosis records missing domain checks as an \textit{Active} cause. (c) Cross-state stitching guides question generation. (d) Later responses verify candidates, and higher self-consistency promotes the cause to \textit{Mastered}.}
  \label{fig:case_study}
\end{figure*}

Figure~\ref{fig:case_study} traces a case on Qwen3-8B-Base. Two solver trajectories for $\log_2(x-1)+\log_2(x-5)=3$ derive $x=3\pm2\sqrt3$. Only the trajectory agreeing with the pseudo-label checks $x>5$. The diagnostician identifies the error cause as missing domain checks after algebraic transformation. It stores this cause as \textit{Active} under the skill node \textit{Checking Conditions after Algebraic Steps}.

In the next round, cross-state stitching pairs this cause with a \textit{Mastered} cause from the same skill node: applying AM-GM without checking whether equality is possible. The resulting question asks for $\inf F+N$, where $F(x)=t+1/t$ and $N$ counts the inputs attaining the infimum. Here $t=\sqrt{(x-2)/(x+1)}$ with $x>2$, so $0<t<1$ rules out AM-GM equality at $t=1$. This tests the original checking rule in a new mathematical form. The question passed double-confidence filtering and entered solver training in round 4.

A later targeted question asks for the sum of solutions to $\sqrt{x+6}=x$. An earlier response accepts both $3$ and $-2$, returning $1$. A later response checks the original equation, rejects $-2$, and returns $3$. Mean self-consistency on questions targeting this cause rises from $48.2\%$ to $65.6\%$ and then $75.0\%$, exceeding the $70\%$ promotion threshold. The cause becomes \textit{Mastered}, ending direct targeting while remaining available for cross-state stitching. The trace illustrates how self-play failures guide what to practice and how memory updates adjust later targeting. Appendix~\ref{sec:appendix_cases} provides the full trace.

\section{Related Work}

\paragraph{Self-Play.}
Self-play trains a challenger and solver on generated tasks \citep{ref8,ref9,xia2026agent0}. Some methods guide generation with human examples or documents \citep{ref16,ref11,ref18}. SESA retrieves skills for its solver, while SPADE uses past environments to guide generation \citep{fu2026sesa,liu2026spade}.

\paragraph{Memory and Curriculum Learning.}
Memory supports task solving \citep{ref38,ref41} and policy training \citep{ref43}. TRACE targets weaknesses found on existing tasks \citep{kang2026trace}. DiagEvo tracks recurring error causes as \textit{Active} or \textit{Mastered}. Their states and recurrence counts guide question generation without external task resources.

\paragraph{Label-Free Reinforcement Learning.}
Label-free RL uses answer agreement \citep{ref8}, reasoning consistency \citep{ref25}, or model confidence \citep{ref26} as reward signals. EMPO filters questions by semantic entropy \citep{zhang2025empo}. Co-rewarding uses question variants or a moving-average teacher, while Self-Harmony combines votes across question variants \citep{zhang2026corewarding,wang2026selfharmony}. DiagEvo's double-confidence filtering retains intermediate-difficulty questions only when the most common answer has a clear vote lead over the second. Appendix~\ref{sec:appendix_related_work} provides additional comparisons.

\section{Conclusion}

DiagEvo turns recurring solver failures into guidance for self-play without external task resources. Its error-cause memory uses cause states and frequencies to balance cause-targeted generation with free exploration. Double-confidence filtering selects training questions. With the default 4B diagnostician, DiagEvo ranks first overall among the compared methods on all three solvers. Qwen3-8B reaches a 72.3\% mathematical reasoning average, 4.5 points above R-Zero. Ablations support the contributions of memory guidance, adaptive scheduling, and filtering.

\clearpage
\bibliographystyle{references}
\bibliography{diagevo}

\clearpage
\appendix

\section{Extended Related Work}
\label{sec:appendix_related_work}

\subsection{Self-Play}

Self-play couples task generation with solver updates. R-Zero generates mathematical questions and trains its challenger with solver uncertainty and a repetition penalty \citep{ref8}. Absolute Zero generates program-based tasks and uses code execution to assess learnability and solution correctness \citep{ref9}. Agent0 adds tool use to this loop. Its curriculum reward combines solver uncertainty, tool calls, and repetition penalties. The executor trains with majority-vote pseudo-labels \citep{xia2026agent0}.

Other methods use external task resources to guide generation. R-Few uses labeled examples both to guide generation and to train the solver \citep{ref16}. SPICE generates questions and reference answers from documents \citep{ref11}. DARC trains its questioner with explicit difficulty targets, then builds an offline curriculum. Its teacher shares the student's parameters but also sees the source documents \citep{ref18}.

Recent methods also retain experience across self-play rounds. SESA distills failed search trajectories into skills retrieved by the solver. This changes solver success rates and thereby the challenger's generation rewards \citep{fu2026sesa}. SPADE stores past executable environments, skill tags, and regret scores to guide environment generation. Its regret signal measures the return gap with and without a hint, while external corpora supply task content \citep{liu2026spade}. DiagEvo stores recurring error causes and directly conditions new questions on them. Cause states and recurrence counts determine which weaknesses to revisit and when to favor free exploration.

\subsection{Memory and Curriculum Learning}

Experience memory can guide later decisions without changing model weights. Reflexion records feedback for later attempts, while ExpeL compares successful and failed trajectories to revise a shared list of insights \citep{ref38,ref39}. AutoGuide compares trajectories with different rewards and extracts instructions for their shared context before the actions diverge \citep{fu2024autoguide}. CER distills environment dynamics and decision skills, then retrieves and updates this memory during interaction \citep{ref40}. ReasoningBank extracts strategies from both successful and failed trajectories \citep{ref41}. ReMe refines procedural memory using retrieval frequency and downstream success \citep{ref42}. ACE maintains a context playbook through reflection and local updates \citep{zhang2026ace}.

Memory can also evolve alongside model parameters. SkillRL couples a skill library with policy training and revises skills in response to validation failures \citep{ref43}. EvolveR alternates experience distillation with policy training. Retrieved principles guide actions, while use and success counts guide memory updates \citep{wu2025evolver}. These methods help the policy use past experience while solving tasks. DiagEvo uses error-cause memory to guide the challenger's question generation.

Curriculum learning controls which examples a model encounters and in what order \citep{ref44}. Online difficulty filtering selects intermediate-difficulty questions from an existing pool using the current policy's success rate \citep{ref45}. SvS synthesizes variants of labeled questions from correct solutions while preserving their answers \citep{ref12}. TRACE compares successful and failed trajectories in target environments to identify missing capabilities. It generates verifiable training environments for these capabilities and combines separately trained adapters \citep{kang2026trace}. DiagEvo extracts error causes from self-play trajectories and tracks them as \textit{Active} or \textit{Mastered} across rounds. These states guide question generation without external task resources.

\subsection{Label-Free Reinforcement Learning}

Label-free RL derives supervision from model outputs. R-Zero uses majority-vote pseudo-labels for generated questions \citep{ref8}, while TTRL applies the same reward principle to adaptation on supplied test questions \citep{zuo2025ttrl}. CoVo combines consistency and volatility in how intermediate reasoning states support candidate answers \citep{ref25}. RLSC instead weights responses by model probabilities, without extracting majority-vote labels \citep{ref26}.

Several methods improve the reliability of these internal signals. EMPO groups answers by meaning and rewards responses according to their group's frequency. It filters questions using lower and upper semantic-entropy thresholds \citep{zhang2025empo}. SRT updates majority-vote pseudo-labels during training and reports gains followed by collapse in longer runs. Its consistency-based filtering delays collapse in the tested setting \citep{shafayat2025srt}. Co-rewarding uses agreement across question variants, a moving-average teacher, or both to stabilize supervision \citep{zhang2026corewarding}. Self-Harmony selects pseudo-labels using the harmonic mean of answer frequencies for original questions and their variants \citep{wang2026selfharmony}. DiagEvo keeps majority-vote supervision. It filters questions by difficulty and the vote ratio between the two most common answers. These checks address vote ambiguity in questions generated by the evolving curriculum.

\section{Case Study: From Failure to Curriculum}
\label{sec:appendix_cases}

This appendix traces one error cause across rounds (Figure~\ref{fig:case_study}), following it through diagnosis, cause-targeted generation, and the subsequent state transition.

\paragraph{Round $t$: failure and diagnosis.}
For $\log_2(x-1)+\log_2(x-5)=3$, both trajectories derive $x=3\pm2\sqrt3$, but only the trajectory agreeing with the pseudo-label enforces $x>5$. The diagnostician records the transferable cause \textit{``Fails to check whether candidate solutions remain valid under the original domain restrictions after algebraic transformation.''} It enters memory as \textit{Active} under the skill node \textit{Checking Conditions after Algebraic Steps}, alongside the \textit{Mastered} cause \textit{``Applies AM-GM to derive a lower bound but does not check whether an allowed input satisfies the equality condition.''}

\paragraph{Round $t+1$: cause-targeted generation and training.}
The challenger samples this \textit{Active} cause and combines it with the related \textit{Mastered} sibling from the same skill node to generate
\[
\begin{aligned}
F(x)={}&\sqrt{\frac{x-2}{x+1}}
       +\sqrt{\frac{x+1}{x-2}},
       \qquad x>2.
\end{aligned}
\]
Let $N$ be the number of real $x>2$ at which $F$ attains its infimum. Find $\inf F+N$. The question requires propagating $0<t<1$ for $t=\sqrt{(x-2)/(x+1)}$ and recognizing that equality at $t=1$ is impossible. It tests the same corrective action under a new surface form. The question passed double-confidence filtering and entered solver training in round 4.

\paragraph{Later rounds: behavior and state.}
A later targeted question asks for the sum of all real solutions to $\sqrt{x+6}=x$. Squaring gives $(x-3)(x+2)=0$. A solver response from an earlier round returns $3+(-2)=1$ without checking the candidates; a response from a later round substitutes them into the original equation, rejects $-2$ because $\sqrt{-2+6}=2\neq-2$, and returns $3$. Across rounds, mean self-consistency on questions targeting this cause rises from $48.2\%$ to $65.6\%$ and then $75.0\%$. The final value exceeds $\theta_{\mathrm{up}}=70\%$, promoting the cause to \textit{Mastered}. The trace illustrates how a diagnosed failure changes the curriculum and later solver behavior.

\section{Limitations and Future Work}

\paragraph{Pseudo-label confidence.}
Double-confidence filtering retains questions whose majority vote share lies within the chosen range and whose leading answer is clearly ahead of the second answer. Both conditions measure agreement among solver responses rather than ground-truth correctness. A shared error can therefore receive strong agreement and pass the filter. Future work can study other verification signals produced during self-play to identify such cases.

\paragraph{Training horizon.}
DiagEvo currently sets the number of co-evolution rounds in advance, while its memory controls question selection within each round. In the seven-round analysis, the mathematical average is highest at round 5. Future work can use changes in the set of \textit{Active} causes and solver self-consistency to decide whether another round is useful.

\paragraph{Evaluation scope.}
DiagEvo builds its curriculum from mathematical questions. The gains on general-reasoning benchmarks show transfer from mathematical training, but do not test direct curriculum construction in other domains. Future work can extend the diagnosis and memory loop to open-ended tasks with long interaction sequences, including multi-turn tool use \citep{ref24}.

\section{Challenger Update and Quality Reward}
\label{sec:appendix_reward}

Following R-Zero \citep{ref8}, each round uses one batch to update the challenger and another to build the solver training set. The current challenger $\theta_t$ first samples $\mathcal{Q}^{\mathrm{chal}}_{t+1}\sim q_{t+1}(\cdot\mid\mathcal{M}_t;\theta_t)$. The frozen solver answers this challenger-update batch. Group Relative Policy Optimization (GRPO) then updates the challenger to $\theta_{t+1}$. The updated challenger samples a new pool of curriculum candidates $\mathcal{Q}^{\mathrm{cur}}_{t+1}\sim q_{t+1}(\cdot\mid\mathcal{M}_t;\theta_{t+1})$. Section~\ref{sec:solver} uses this pool to construct and filter the solver training set.

For a format-valid question $x$, let $p_1(x)$ denote the frozen solver's majority-answer vote share. The uncertainty reward is
\begin{equation}
  r_{\mathrm{unc}}(x)=1-2\left|p_1(x)-\frac{1}{2}\right|.
  \label{eq:uncertainty_reward}
\end{equation}
It peaks at 50\% self-consistency, favoring questions near the solver's competence boundary.

The repetition penalty in Equation~\ref{eq:challenger_reward} follows R-Zero \citep{ref8}. For the challenger-update batch $\mathcal{X}=\{x_i\}_{i=1}^{B}$, we compute the pairwise distance
\begin{equation}
  d_{ij}=1-\operatorname{BLEU}(x_i,x_j).
\end{equation}
Agglomerative clustering with the R-Zero criterion produces clusters $\mathcal{C}=\{C_1,\ldots,C_K\}$. For $x_i\in C_k$, the repetition penalty is $r_{\mathrm{rep}}(x_i)=|C_k|/B$, so large clusters receive larger penalties.

The full reward is
\begin{equation}
  r_{\mathrm{chal}}(x)
  =\max\!\bigl\{0,\;r_{\mathrm{unc}}(x)-\lambda r_{\mathrm{rep}}(x)\bigr\},
  \label{eq:challenger_reward}
\end{equation}
with $\lambda=1$ following R-Zero; format-invalid questions receive zero reward.

\section{Training Hyperparameters}
\label{sec:appendix_hyperparams}

This section reports the settings used to reproduce DiagEvo. We first list the GRPO configuration shared by the challenger and solver, then report the response-group and co-evolution settings, followed by the reward, filtering, and curriculum thresholds.

\paragraph{GRPO Optimization.}
Both the challenger and solver use the learning rate, weight decay, KL penalty coefficient, rollout temperature, and top-$p$ reported by R-Zero, as summarized in Table~\ref{tab:appendix_grpo}. Their role-specific batch sizes, response-group sizes, and update settings are reported separately in Table~\ref{tab:appendix_diagevo}. The surrogate objective uses the default clip range in the standard implementation. All experiments run on a single node of 8 NVIDIA H200 GPUs with BF16 mixed precision and FlashAttention 2.

\begin{table}[H]
  \caption{GRPO optimization settings for the challenger and solver.}
  \label{tab:appendix_grpo}
  \centering
  \begin{tabular}{lc}
    \toprule
    Setting & Value \\
    \midrule
    Learning rate & $1 \times 10^{-6}$ \\
    Weight decay & $1 \times 10^{-2}$ \\
    KL penalty coefficient $\beta$ & $1 \times 10^{-2}$ \\
    Rollout temperature & 1.0 \\
    Rollout top-$p$ & 0.99 \\
    \bottomrule
  \end{tabular}
\end{table}

\paragraph{Batch Size and Response Groups.}
In each round, the challenger is first optimized for up to 6 GRPO steps. Each challenger step uses a global batch size of 256 and a group size of eight. The updated challenger then generates the curriculum candidates. For every candidate, the current solver samples 12 construction responses to determine the pseudo-label, $p_1$, and $p_2$. For each \textit{Active} cause, the $p_1$ values of all candidates targeting that cause are averaged to obtain the self-consistency score used for promotion. Candidates whose construction votes satisfy both confidence constraints form the solver training set. For each retained question, one construction response agreeing with its pseudo-label is stored as the reference trajectory for diagnosis. The solver is then optimized for up to 12 GRPO steps. Each solver step uses a global batch size of 256. For each retained question in the batch, the solver samples a separate group of eight optimization responses. These responses provide the policy-gradient signal, while their failed trajectories are paired with the stored reference for diagnosis and memory maintenance. Table~\ref{tab:appendix_diagevo} summarizes these settings.

\begin{table}[H]
  \caption{DiagEvo-specific training configuration (challenger and solver co-evolution loop).}
  \label{tab:appendix_diagevo}
  \centering
  \begin{tabular}{lc}
    \toprule
    Hyperparameter & Value \\
    \midrule
    Solver GRPO global batch size & 256 \\
    Challenger GRPO global batch size & 256 \\
    Solver GRPO group size $G$ & 8 \\
    Challenger GRPO group size & 8 \\
    Solver max steps per round & 12 \\
    Challenger max steps per round & 6 \\
    Construction responses per candidate $N$ & 12 \\
    Main-experiment training rounds & 5 \\
    \bottomrule
  \end{tabular}
\end{table}

\paragraph{Reward, filtering, curriculum, and memory-maintenance hyperparameters.}
Table~\ref{tab:appendix_thresholds} reports the numerical settings that control the challenger reward, double-confidence filtering, curriculum transitions, and memory maintenance. Appendix~\ref{sec:appendix_prompts} specifies the diagnostician routing and retrieval procedure.

\begin{table}[H]
  \caption{Reward, filtering, curriculum, and memory-maintenance hyperparameters used throughout DiagEvo.}
  \label{tab:appendix_thresholds}
  \centering
  \small
  \setlength{\tabcolsep}{4pt}
  \begin{tabular}{lc}
    \toprule
    Symbol & Value \\
    \midrule
    Repetition penalty weight $\lambda$ & 1.00 \\
    Absolute difficulty lower bound $p_{\mathrm{low}}$ & 0.25 \\
    Absolute difficulty upper bound $p_{\mathrm{high}}$ & 0.75 \\
    Relative confidence threshold $\tau$ & 1.60 \\
    State promotion threshold $\theta_{\mathrm{up}}$ & 0.70 \\
    Exploration-exploitation proportionality $k$ & 0.50 \\
    Deduplication similarity threshold $\theta_{\mathrm{dup}}$ & 0.50 \\
    Node-merge similarity threshold $\theta_{\mathrm{merge}}$ & 0.50 \\
    Node candidate shortlist size $K$ & 5 \\
    \bottomrule
  \end{tabular}
\end{table}

\paragraph{Implementation Details.}
All three Qwen3-Instruct-2507 diagnosticians run in non-thinking mode. Qwen3-4B-Instruct-2507 is the default for experiments that do not compare diagnostician scale. For questions retained by double-confidence filtering, the diagnostician compares failed optimization trajectories with stored construction references. It reads only questions, pseudo-labels, and trajectories produced during self-play. The embedding model reads only error causes derived from these trajectories. It uses text-embedding-v4 for error-cause retrieval. Both models are frozen and general-purpose. Section~\ref{sec:memory} describes how these models are used for extraction and deduplication.

\paragraph{Training Time Breakdown.}
Table~\ref{tab:appendix_runtime} reports each component's share of total elapsed time in a complete DiagEvo training run. Question sampling, pseudo-label construction, and the GRPO updates of the two models account for 94.2\% of the total time; each of these components has a direct counterpart in R-Zero. Diagnosis and memory maintenance account for 5.7\% of total training time, with 3.9\% spent on comparative error-cause extraction.

\begin{table}[H]
  \caption{Training time breakdown for a complete DiagEvo run with the default configuration. Diagnosis and memory maintenance, the components specific to DiagEvo, account for 5.7\% of the total time.}
  \label{tab:appendix_runtime}
  \centering
  \small
  \setlength{\tabcolsep}{4pt}
  \begin{tabular}{p{0.72\columnwidth}r}
    \toprule
    Component & Training time (\%) \\
    \midrule
    Challenger policy optimization & 27.1 \\
    Curriculum candidate generation & 10.9 \\
    Pseudo-label construction and double-confidence filtering & 30.3 \\
    Solver policy optimization & 25.9 \\
    \textbf{Diagnosis and memory maintenance} & \textbf{5.7} \\
    \quad Comparative error-cause extraction & 3.9 \\
    \quad Deduplication and node assignment & 1.3 \\
    \quad Skill-node consolidation & 0.4 \\
    \quad State--frequency update and scheduling & 0.1 \\
    Other orchestration & 0.1 \\
    \bottomrule
  \end{tabular}
\end{table}

\section{Evaluation Protocol and Baseline Score Validation}
\label{sec:appendix_repro}

\paragraph{Validation and Checkpoint Selection.}
The fixed validation set contains 600 questions. We sample 300 Level 3 and Level 4 questions from the MATH training split \citep{ref29}, covering all seven subject categories. Another 300 questions come from the MMLU validation split \citep{hendrycks2021mmlu}, covering all 57 subjects. These questions are disjoint from the nine test benchmarks. We tune hyperparameters using accuracy over all 600 validation questions and fix the defaults before component ablations. Each run uses the checkpoint with the highest validation accuracy for evaluation on all nine test benchmarks.

\paragraph{Benchmark Evaluation.}
For mathematical reasoning, we follow the simple-evals protocol and use GPT-4o as an automatic judge, matching DARC \citep{ref18}. AMC contains 40 questions. Following R-Zero \citep{ref8}, we sample 32 responses per AMC question and report their mean accuracy (mean@32). For general reasoning, we use exact-match accuracy with greedy decoding.

\paragraph{Baseline Score Validation.}
Four baseline rows in Table~\ref{tab:main_results} are taken from DARC \citep{ref18}. To verify that these scores are comparable to our own runs, we re-evaluate the three base models and reproduce R-Zero \citep{ref8} under our own infrastructure. We use the same solvers, benchmarks, and evaluation protocol. Table~\ref{tab:appendix_repro} compares the aggregate scores. Base-model averages differ by at most 0.1 points, and R-Zero scores differ by at most 0.5 points. Per-benchmark deviations remain below 1.0 points. Table~\ref{tab:main_results} therefore reports our own scores for the base models and R-Zero. It reports the scores from DARC for the remaining baselines. DARC does not report R-Few for OctoThinker-8B-Hybrid-Base, so Table~\ref{tab:main_results} omits this setting.

\begin{table}[htbp]
  \caption{Aggregate scores of the reported baselines and of our reproduction under the same evaluation protocol. Base-model averages match within 0.1 points and R-Zero scores within 0.5 points on every column.}
  \label{tab:appendix_repro}
  \centering
  \begin{tabular}{llccc}
    \toprule
    Solver & Scores & Math Avg & Gen Avg & Overall \\
    \midrule
    \multirow{4}{*}{Qwen3-4B-Base}
      & Base (reported)     & 53.1 & 27.9 & 41.9 \\
      & Base (reproduced)   & 53.0 & 27.8 & 41.8 \\
      & R-Zero (reported)   & 61.1 & 32.2 & 48.2 \\
      & R-Zero (reproduced) & 60.6 & 31.9 & 47.9 \\
    \midrule
    \multirow{4}{*}{Qwen3-8B-Base}
      & Base (reported)     & 63.3 & 33.1 & 49.9 \\
      & Base (reproduced)   & 63.4 & 33.2 & 50.0 \\
      & R-Zero (reported)   & 67.6 & 36.3 & 53.7 \\
      & R-Zero (reproduced) & 67.8 & 36.2 & 53.8 \\
    \midrule
    \multirow{4}{*}{OctoThinker-8B-Hybrid-Base}
      & Base (reported)     & 35.8 & 10.6 & 24.6 \\
      & Base (reproduced)   & 35.9 & 10.7 & 24.7 \\
      & R-Zero (reported)   & 46.4 & 21.2 & 35.2 \\
      & R-Zero (reproduced) & 46.0 & 20.9 & 34.9 \\
    \bottomrule
  \end{tabular}
\end{table}

\section{Generation Schedule and Hyperparameter Sensitivity}
\label{sec:appendix_sensitivity}

This appendix compares fixed and adaptive generation and examines sensitivity to $k$ (Equation~\ref{eq:eps}) and $\tau$ (Equation~\ref{eq:filtering}). All runs use Qwen3-8B-Base with the default 4B diagnostician.

\paragraph{Fixed and Adaptive Generation.}
We compare the frequency-driven schedule with fixed ratios of free exploration to cause-targeted generation, keeping all other settings unchanged (Table~\ref{tab:ablation_fixed_ratio}). The schedule achieves higher mathematical and general averages than all three fixed ratios.

\begin{table}[htbp]
  \centering
  \caption{Fixed and adaptive generation. Ratios denote free exploration to cause-targeted generation.}
  \label{tab:ablation_fixed_ratio}
  \begin{tabular}{lcc}
    \toprule
    Generation policy & Math Avg. & General Avg. \\
    \midrule
    Fixed 7:3 & 70.1 & 36.9 \\
    Fixed 5:5 & 70.9 & 37.2 \\
    Fixed 3:7 & 71.2 & 37.6 \\
    \midrule
    Frequency-driven (full) & \textbf{72.3} & \textbf{38.8} \\
    \bottomrule
  \end{tabular}
\end{table}

\paragraph{Hyperparameter Sensitivity.}

The default values, $k=0.5$ and $\tau=1.6$, are selected on the validation set. Each sensitivity experiment varies one of them while keeping other settings fixed and reports test benchmark scores.

Table~\ref{tab:ablation_k} varies $k$ in the frequency-driven schedule. Smaller values favor earlier exploitation, whereas larger values preserve more exploration.

\begin{table}[H]
  \centering
  \caption{Sensitivity of the frequency-driven schedule to $k$ (Qwen3-8B-Base).}
  \label{tab:ablation_k}
  \begin{tabular}{lcc}
    \toprule
    $k$ & Math Avg & General Avg \\
    \midrule
    0.2 & 71.4 & 38.1 \\
    0.5 (default) & \textbf{72.3} & \textbf{38.8} \\
    1.0 & 71.8 & 38.4 \\
    2.0 & 71.3 & 37.8 \\
    \bottomrule
  \end{tabular}
\end{table}

Table~\ref{tab:ablation_tau} varies $\tau$ in double-confidence filtering; the $\tau=1.6$ row repeats the full configuration from Table~\ref{tab:ablation} for reference. Performance falls on either side of $\tau=1.6$. The drop under stricter filtering is consistent with removing too many training questions, although we do not directly measure retention rates.

\begin{table}[H]
  \centering
  \caption{Sensitivity of double-confidence filtering to the relative confidence threshold $\tau$ (Qwen3-8B-Base).}
  \label{tab:ablation_tau}
  \begin{tabular}{lcc}
    \toprule
    $\tau$ & Math Avg & General Avg \\
    \midrule
    1.2 (loose) & 71.6 & 38.4 \\
    1.6 (default) & \textbf{72.3} & \textbf{38.8} \\
    2.0 (strict) & 71.3 & 38.1 \\
    \bottomrule
  \end{tabular}
\end{table}

\section{Diagnostic Evaluation Protocols}
\label{sec:appendix_analysis}

\begin{figure*}[!t]
  \centering
  \begin{subfigure}{0.32\textwidth}
    \centering
    \includegraphics[width=\linewidth]{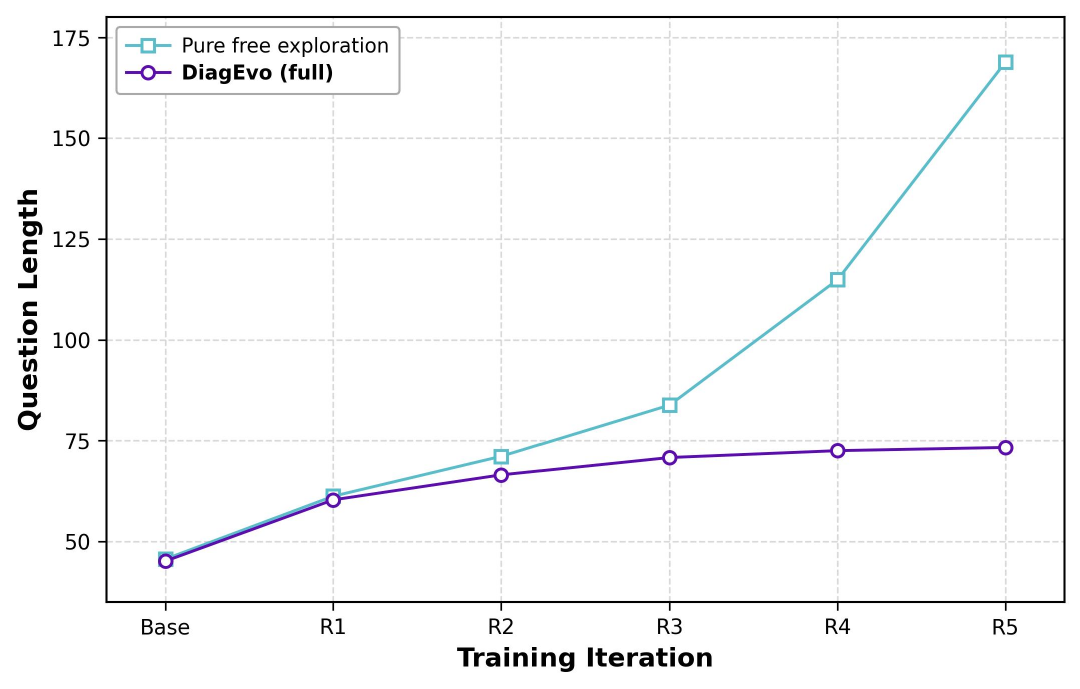}
    \caption{Question length}
    \label{fig:props_length}
  \end{subfigure}\hfill
  \begin{subfigure}{0.32\textwidth}
    \centering
    \includegraphics[width=\linewidth]{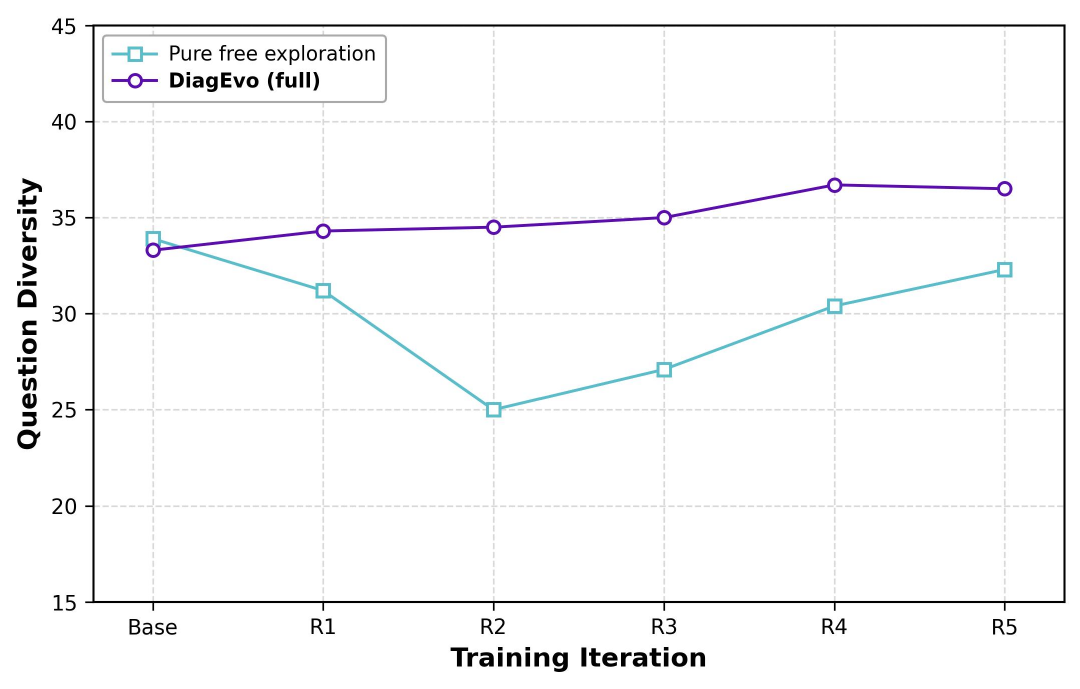}
    \caption{2-gram diversity}
    \label{fig:props_diversity}
  \end{subfigure}\hfill
  \begin{subfigure}{0.32\textwidth}
    \centering
    \includegraphics[width=\linewidth]{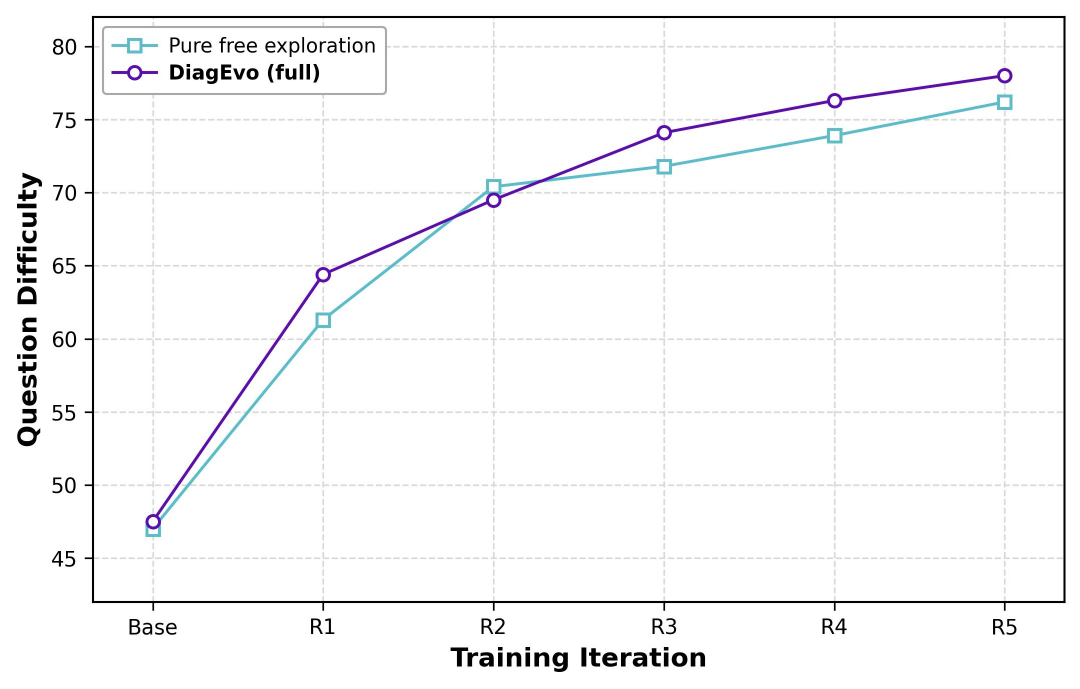}
    \caption{Question difficulty}
    \label{fig:props_difficulty}
  \end{subfigure}
  \caption{Properties of questions generated after each round on Qwen3-8B-Base. (a) Average length in words. (b) Lexical diversity over adjacent word pairs, reported as 2-gram diversity. (c) Difficulty measured by the error rate of a fixed Qwen3-8B-Base solver against oracle answers. Both variants use double-confidence filtering. \emph{Pure free exploration} removes the error-cause memory.}
  \label{fig:question_properties}
\end{figure*}

\begin{figure*}[!t]
  \centering
  \begin{subfigure}{0.32\textwidth}
    \centering
    \includegraphics[width=\linewidth]{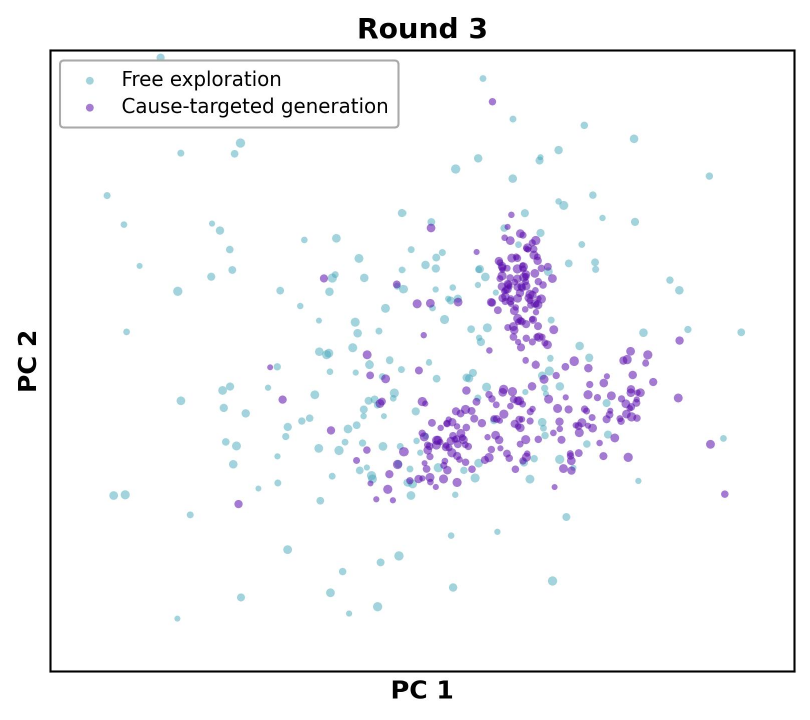}
    \caption{Question coverage, round 3}
    \label{fig:coverage_r3}
  \end{subfigure}\hfill
  \begin{subfigure}{0.32\textwidth}
    \centering
    \includegraphics[width=\linewidth]{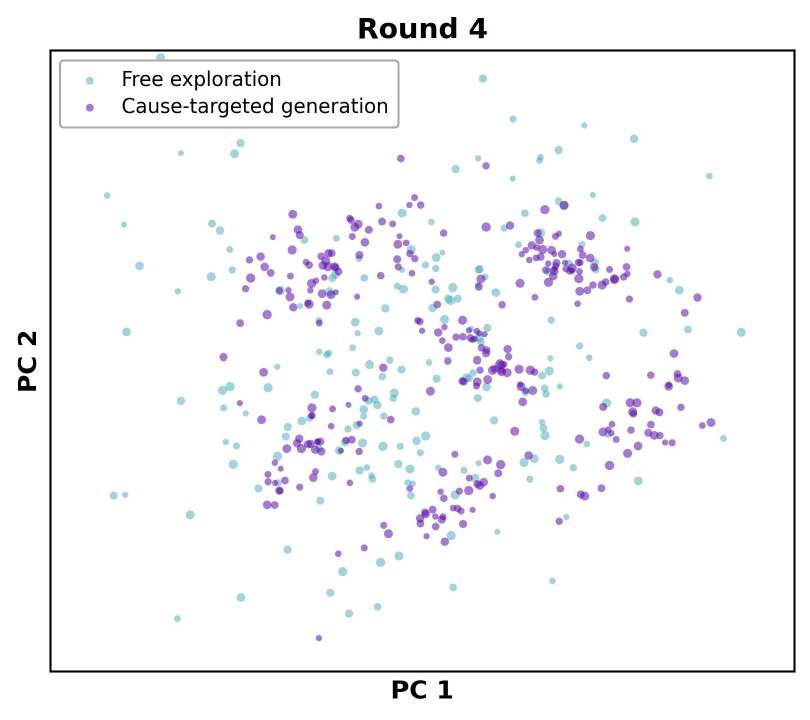}
    \caption{Question coverage, round 4}
    \label{fig:coverage_r4}
  \end{subfigure}\hfill
  \begin{subfigure}{0.32\textwidth}
    \centering
    \includegraphics[width=\linewidth]{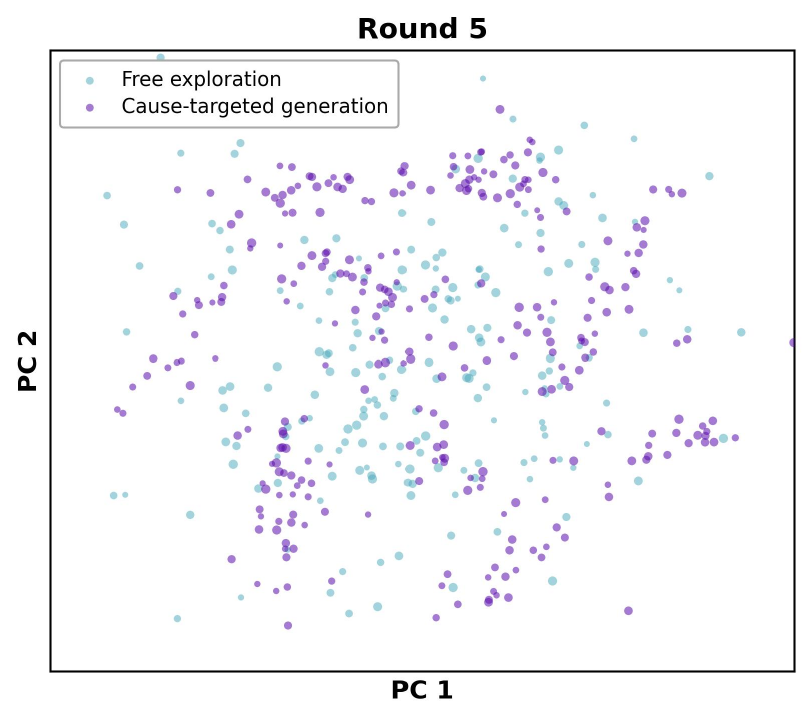}
    \caption{Question coverage, round 5}
    \label{fig:coverage_r5}
  \end{subfigure}
  \caption{Principal component analysis (PCA) projections of questions generated in rounds 3--5. Colors indicate the generation mode.}
  \label{fig:coverage}
\end{figure*}

This section specifies the measurements behind Figures~\ref{fig:dynamics}, \ref{fig:question_properties}, and~\ref{fig:coverage}. All of them use Qwen3-8B-Base with the default 4B diagnostician, and every variant shares the training configuration in Appendix~\ref{sec:appendix_hyperparams}.

\paragraph{Per-round performance.}
After each round, we evaluate the current solver under the protocol in Appendix~\ref{sec:appendix_repro} and report the mathematical reasoning average. Both curves in Figure~\ref{fig:dynamics}(a) retain double-confidence filtering and differ only in whether question generation uses the error-cause memory. We additionally evaluate both variants in rounds 6 and 7 under the same protocol; Section~\ref{sec:dynamics} reports the resulting averages.

\paragraph{Memory statistics.}
We count the \textit{Active} and \textit{Mastered} error causes and the skill nodes in the memory committed after each round (Equation~\ref{eq:memory_representation}).

\paragraph{Pseudo-label accuracy.}
We track pseudo-label accuracy across rounds, as in R-Zero \citep{ref8}. In each round, we select 200 challenger-generated questions and use the same set for all three variants. Each variant is evaluated using its solver checkpoint after that round's training. For every question, the solver samples 12 responses and takes their majority answer as the pseudo-label, matching the response count for curriculum construction. Qwen3.8-Max proposes each reference answer by majority vote over 16 sampled responses. Humans verify every answer and correct any errors to obtain the oracle answers, which are used only for evaluation. We report the percentage of each variant's majority-vote pseudo-labels that match these oracle answers across all 200 questions.

\paragraph{Question properties.}
Following the protocol of R-Few \citep{ref16}, we measure the average question length in words and the 2-gram lexical diversity of the questions generated after each round. Difficulty is the error rate of a fixed Qwen3-8B-Base solver on these questions, with answers relabeled by Qwen3.8-Max; since this oracle differs from the one used by R-Few, the absolute values should not be compared across papers. As in Figure~\ref{fig:dynamics}(a), both variants retain double-confidence filtering.

Figure~\ref{fig:question_properties} tests whether question difficulty grows with lexical diversity or simply with question length. Without the memory, average length grows from 45.6 to 168.9 words. The 2-gram diversity rises again only after the questions become much longer. This timing links the late increase in 2-gram diversity to the added length.

With the memory, average length stays near 73 words after round 2. At the same time, 2-gram diversity rises from 33.3 to 36.5. Both variants generate harder questions over the rounds. The memory-guided variant does so at a stable length. Its increase in difficulty therefore cannot be explained by longer questions.

\paragraph{Coverage visualization.}
We embed the questions generated after each round with text-embedding-v4 and project the embeddings to two dimensions with PCA. Figure~\ref{fig:coverage} colors each question by its generation mode, free exploration or cause-targeted.

The changing memory also changes what the challenger generates (Figure~\ref{fig:coverage}). Free exploration covers a broad region that remains stable across rounds. Cause-targeted generation moves into new regions as the memory gains error causes. The two modes therefore have complementary roles. Free exploration maintains broad coverage, while cause-targeted generation follows the solver's current error causes. This pattern is consistent with the performance loss when Table~\ref{tab:ablation} removes either mode.

\section{Prompt Templates and Memory Routing}
\label{sec:appendix_prompts}

This section specifies the runtime prompts and routing procedures for the challenger, solver, and diagnostician. Challenger and solver prompts follow the level of detail in R-Zero \citep{ref8}; diagnostician prompts implement the semantic decisions described in Section~\ref{sec:memory}. Fields marked \texttt{<xxx>} are replaced at runtime.

\paragraph{Design rationale.}
Pseudo-labels come from solver majority voting (Section~\ref{sec:solver}), so the challenger does not provide answers. We retain the role, difficulty criterion, and \texttt{<question>} format of R-Zero \citep{ref8}, but omit its \texttt{\textbackslash boxed\{final answer\}} requirement because DiagEvo constructs pseudo-labels from solver votes. Diagnostic context appears only in the challenger user message: either one error cause or an \textit{Active}/\textit{Mastered} pair. The shared system message is unchanged across generation modes. Each cause is accompanied by its parent skill-node label to provide broader knowledge context.

\paragraph{Shared System Message.}
The following system message is used for all three generation modes described below.
\begin{quote}
You are an expert competition-math problem setter. FIRST, in your private scratch-pad, think step-by-step to design a brand-new, non-trivial problem. The problem could come from any field of mathematics, including but not limited to algebra, geometry, number theory, combinatorics, prealgebra, probability, statistics, and calculus. Aim for a difficulty such that fewer than 30\% of advanced high-school students could solve it. Avoid re-using textbook clich\'es or famous contest problems.

THEN, without revealing any of your private thoughts, output \textit{exactly} the following block:

\texttt{<question>}

\{The full problem statement on one or more lines\}

\texttt{</question>}

Do NOT output anything else---no explanations, no extra markup.
\end{quote}

\paragraph{Mode 1: Free Exploration.}
The challenger relies solely on its own internal knowledge, with no reference to the memory (Section~\ref{sec:generation}).
\begin{quote}
\textbf{User Message:}
Generate one new, challenging reasoning question now. Remember to format the output exactly as instructed.
\end{quote}

\paragraph{Mode 2: Cause-Targeted Generation (Single Cause).}
Activated when the sampled \textit{Active} error cause's skill node contains no \textit{Mastered} causes (Section~\ref{sec:generation}).
\begin{quote}
\textbf{User Message:}
Generate one new, challenging reasoning question now. The question must specifically target the following reasoning weakness observed in the solver's failures:

Knowledge area: \texttt{<skill\_node>}

Weakness to probe: \texttt{<error\_cause>}

Design the question so that correctly solving it requires the solver to overcome this specific weakness. Remember to format the output exactly as instructed.
\end{quote}

\paragraph{Mode 3: Cross-State Stitching.}
Activated when the sampled \textit{Active} error cause's skill node also contains at least one \textit{Mastered} cause (Section~\ref{sec:generation}); the two causes are combined in a single composite question rather than presented as separate tasks.
\begin{quote}
\textbf{User Message:}
Generate one new, challenging reasoning question now. The question must weave together two reasoning elements within the same knowledge area into a single, coherent problem, rather than testing them separately:

Knowledge area: \texttt{<skill\_node>}

Element A (an observed weakness that remains \textit{Active} and should still be challenging): \texttt{<active\_cause>}

Element B (a previously diagnosed weakness for which the solver has reached high self-consistency, as measured by the endogenous consistency proxy; the question should require the corresponding corrective reasoning skill): \texttt{<mastered\_cause>}

Design a single self-contained question that genuinely requires combining both elements to reach the answer---neither element should be solvable in isolation from the other. Remember to format the output exactly as instructed.
\end{quote}

\paragraph{Solver Prompt.}
The solver prompt is identical to that of R-Zero \citep{ref8} because answer extraction is independent of question generation. R-Few \citep{ref16} uses the same instruction, while SPICE \citep{ref11} instead uses model-specific reasoner templates.
\begin{quote}
\textbf{System Message:}
Please reason step by step, and put your final answer within \verb|\boxed{}|.

\textbf{User Message:}
\texttt{<question>}
\end{quote}

\paragraph{Diagnostician Prompts.}
The diagnostician operates within a hybrid code-and-LLM pipeline that follows the memory-maintenance sequence in Section~\ref{sec:memory}. Code performs similarity screening, top-$K$ retrieval, and routing; the diagnostician performs extraction, deduplication, node assignment, and consolidation. Recurrence verification is folded into extraction. The diagnostician returns JSON outputs in the runtime order described below. Error causes and node labels must be 10--20-word verb--object phrases without question-specific values or variables, which keeps their embeddings focused on transferable semantics.

\paragraph{Pre-Check: Known-Cause Recurrence Verification (LLM Call \#1, folded into Prompt 1).}
For each question retained by Equation~\ref{eq:filtering}, diagnosis pairs failed optimization trajectories with the stored construction reference that agrees with its pseudo-label. No extra responses are sampled for diagnosis. The known-cause list is empty for free exploration, contains the sampled \textit{Active} cause for single-cause generation, and contains both the sampled \textit{Active} cause and the stitched \textit{Mastered} cause for cross-state generation. Prompt~1 first locates the earliest reasoning difference and compares it with every supplied cause. It assigns each failed trajectory to at most one supplied cause, selecting the single closest match when more than one appears plausible. A confirmed match contributes one count to $\Delta_t(e)$ for the matched cause; otherwise, the prompt extracts a new candidate cause for memory matching. Both paths use the same LLM call.

\paragraph{Prompt 1: Error-Cause Extraction (LLM Call \#1).}
\begin{quote}
\textbf{System Message:}
You are a diagnostician analyzing why a math solver's reasoning went wrong. You will be shown a problem, a working reference answer, and two solving attempts that disagree with each other: one that reaches the reference answer and one that does not. Your job is to locate the exact reasoning step where the two attempts diverge and name the underlying skill deficiency responsible for the divergence.

Do not re-solve the problem yourself, and do not treat the reference answer as ground truth: it was itself produced by majority voting over the solver's own samples and can occasionally be wrong. Your only task is to explain the divergence between the two given attempts, not to adjudicate correctness from outside evidence.

If one or more known causes are supplied below, compare the located divergence with every supplied cause. Select at most one cause whose underlying failure mechanism best explains the earliest divergence. Only if none of the supplied causes matches should you characterize a new one.

\begin{samepage}
EXTRACTION PRINCIPLES:
\begin{itemize}
\item Localize the single earliest step at which the two attempts stop agreeing, and diagnose that step specifically---do not summarize the disagreeing attempt's overall approach.
\item State the error cause as a general, transferable skill deficiency, phrased as a single verb--object phrase (e.g., ``misapplies triangle inequality to algebraic side lengths''), never as a restatement of this problem's specific numbers, entities, or context.
\item The description must be 10--20 words. This is a hard constraint, not a suggestion: it will later be embedded and used for cosine-similarity screening before semantic confirmation, so any leftover problem-specific detail, hedge, or explanation will dilute the embedding and silently corrupt every downstream deduplication and routing decision.
\item If the two attempts do not actually diverge in reasoning strategy (e.g., the disagreement is a trivial arithmetic or copying slip), report this explicitly rather than forcing an artificial skill label onto a non-skill failure.
\end{itemize}
\end{samepage}

\textbf{User Message:}
\# Problem
\texttt{<problem>}

\# Working Reference Answer
\texttt{<pseudo\_label>}

\# Attempt That Agrees With the Reference
\texttt{<agreeing\_response>}

\# Attempt That Disagrees With the Reference
\texttt{<disagreeing\_response>}

\# Known Causes Used for Generation (omit this section if the problem was freely generated)
\texttt{<known\_cause\_list>}

Each entry contains a cause ID, its current state, and its cause description.

OUTPUT FORMAT:
Return exactly one JSON object with the following fields.
\begin{samepage}
\begin{verbatim}
{
  "matched_cause_id": null,
  "error_cause": null
}
\end{verbatim}
\end{samepage}
Set \texttt{matched\_cause\_id} to the exact ID of the single supplied cause that best explains the earliest divergence, or to \texttt{null} if none matches. When \texttt{matched\_cause\_id} is non-null, set \texttt{error\_cause} to \texttt{null}; the pipeline assigns the observation directly to that cause and skips the stages below. When \texttt{matched\_cause\_id} is \texttt{null}, \texttt{error\_cause} must be a new 10--20-word verb--object description unless the failure is unrelated to any identifiable skill, in which case it is also \texttt{null}.
\end{quote}

\paragraph{Stage 2a: Error-Cause Similarity Screening (non-LLM).}
For each new candidate cause, we compare its embedding with every stored cause. Candidates above $\theta_{\mathrm{dup}}$ proceed to Prompt~2 for semantic confirmation. If none qualifies, the candidate proceeds directly to node assignment.

\paragraph{Prompt 2: Deduplication (LLM Call \#2).}
Invoked only when Stage~2a returns at least one candidate.
\begin{quote}
\textbf{System Message:}
You are a diagnostician deciding whether a newly identified error cause describes a mistake already recorded in the memory. You will be shown one new error cause and a short list of existing candidates that an embedding search flagged as textually similar to it. Judge duplication by underlying mechanism, not by wording: two causes represent the same error cause if fixing one would also fix the other, even when phrased differently; they are different if a solver could overcome one while still exhibiting the other.

Your decision directly controls the memory's current-episode frequency signal and may reactivate a matched \textit{Mastered} cause. Frequency determines sampling weights within the Active cause-targeted pool and contributes to the next round's exploration--exploitation ratio: confirming a genuine duplicate adds evidence to an \textit{Active} cause or starts a reactivated cause at one, while wrongly accepting two distinct failure modes as one would blur the signal used for cause-targeted generation. Err toward keeping mechanistically distinct causes apart whenever the match is not clearly correct.

\textbf{User Message:}
\# New Error Cause
\texttt{<new\_cause>}

\# Candidate Existing Causes
\texttt{<candidate\_cause\_list>}

OUTPUT FORMAT:
Return exactly one of the following valid JSON objects.
\begin{samepage}
\begin{verbatim}
{
  "duplicate": true,
  "matched_cause_id": "<matched_cause_id>"
}
\end{verbatim}
\end{samepage}
or
\begin{samepage}
\begin{verbatim}
{
  "duplicate": false,
  "matched_cause_id": null
}
\end{verbatim}
\end{samepage}
If more than one candidate plausibly matches, select the single closest one. When \texttt{duplicate} is \texttt{true}, the observation is assigned to \texttt{matched\_cause\_id} and contributes one count to $\Delta_t(e)$ in Equation~\ref{eq:state_frequency_transition}; no new cause is created. The joint transition then increments an \textit{Active} cause or starts a new active episode for a \textit{Mastered} cause with the matched-failure count as its frequency. When \texttt{duplicate} is \texttt{false}, the new cause proceeds to Stage~3a for node assignment.
\end{quote}

\paragraph{Stage 3a: Node Candidate Retrieval (non-LLM).}
We reuse the cause-level similarities from Stage~2a to retrieve candidate skill nodes. A node receives the maximum similarity between the new cause and any of its members, and the top-$K$ nodes form the shortlist. If the memory has at most $K$ nodes, all nodes are returned.

\paragraph{Prompt 3: Skill Node Assignment (LLM Call \#3).}
Prompt~3 is invoked once for each new error cause that clears deduplication, using the top-$K$ skill-node candidates from Stage~3a.
\begin{quote}
\textbf{System Message:}
You are a diagnostician organizing error causes under skill nodes. Each node names a reasoning or knowledge skill that the challenger can target on its own. You will see a new error cause and its top-$K$ most relevant existing skill nodes. Each node is shown with its label and one representative error cause. Base your decision only on this shortlist; do not assume access to the full memory. In a single pass, decide whether the new cause belongs to a listed node or needs a new one. Use the following three criteria:

\begin{itemize}
\item \textbf{Semantic distinctness:} is the new cause's underlying skill clearly different from every candidate node's skill, as defined relative to this specific candidate set (this criterion is trivially satisfied when no candidates are shown)?
\item \textbf{Independent actionability:} would practicing to fix this cause be a meaningfully different exercise from practicing any candidate node's skill? A node should correspond to one coherent thing the solver can get better at, not a loose bag of unrelated causes.
\item \textbf{Generality and extensibility:} if a new node is created, is its label general enough to later host other, distinct error causes that share this skill, rather than merely restating this one instance? A node opened too narrowly will never accumulate enough error causes to reveal a stable pattern for later composite-problem generation.
\end{itemize}

\textbf{User Message:}
\# New Error Cause
\texttt{<new\_cause>}

\# Top-$K$ Candidate Skill Nodes
\texttt{<candidate\_node\_list>}

OUTPUT FORMAT:
Return exactly one of the following valid JSON objects.
\begin{samepage}
\begin{verbatim}
{
  "create_new_node": true,
  "new_node_label": "<10-20 word verb-object label>",
  "assigned_node": null
}
\end{verbatim}
\end{samepage}
or
\begin{samepage}
\begin{verbatim}
{
  "create_new_node": false,
  "new_node_label": null,
  "assigned_node": "<assigned_node_id>"
}
\end{verbatim}
\end{samepage}
Exactly one of \texttt{new\_node\_label} and \texttt{assigned\_node} must be non-null. This single pass is the only chance to get the decision right: it is trusted and executed as-is, with no second retrieval or second diagnostician call to verify it, so weigh all three criteria within this one judgment rather than treating creation and assignment as separate questions.
\end{quote}

\paragraph{Stage 3.5: Routing Execution (code only, no LLM or embedding call).}
Code executes Prompt~3's routing decision directly. If \texttt{create\_new\_node} is \texttt{true}, it creates a skill node labeled \texttt{new\_node\_label} and attaches the error cause as its first member; otherwise, it attaches the error cause under \texttt{assigned\_node}. Either way, the new error cause is initialized with $(s_t(e),f_t(e))=(\textit{Active},1)$, consistent with Equation~\ref{eq:memory_representation}.

\paragraph{Stage 4a: Node-Pair Candidate Screening (non-LLM).}
After all failures in round $t$ are processed and routed, we compare skill-node embeddings pairwise. Pairs above $\theta_{\mathrm{merge}}$ proceed to Prompt~4 for semantic confirmation.

\paragraph{Prompt 4: End-of-Round Consolidation (LLM Call \#4).}
Prompt~4 is invoked once for each candidate pair from Stage~4a.
\begin{quote}
\textbf{System Message:}
You are a diagnostician checking skill nodes for redundancy as the memory grows. Prompt~3 assigns each error cause using only the shortlist available at that time. Nodes created at different times may therefore describe the same skill. Your task is to identify these duplicate nodes. You will see two skill nodes, each with its label and all its error causes. Decide whether they describe the same skill and should be merged, or distinct skills that share similar words. Base your decision on all error causes under both nodes. Similar labels can cover different mistakes, while different labels can cover the same mistake.

\textbf{User Message:}
\# Skill Node A
\texttt{<node\_a\_label>}: \texttt{<node\_a\_causes>}

\# Skill Node B
\texttt{<node\_b\_label>}: \texttt{<node\_b\_causes>}

OUTPUT FORMAT:
Return exactly one of the following valid JSON objects.
\begin{samepage}
\begin{verbatim}
{
  "merge": true,
  "merged_label": "<10-20 word verb-object label>"
}
\end{verbatim}
\end{samepage}
or
\begin{samepage}
\begin{verbatim}
{
  "merge": false,
  "merged_label": null
}
\end{verbatim}
\end{samepage}
The merged label must remain a general, transferable skill description under the same 10--20 word verb--object phrase constraint as any node label, not a concatenation of A's and B's labels. When \texttt{merge} is \texttt{true}, the pipeline relabels the surviving node as \texttt{merged\_label} and re-parents every error cause from both A and B under it without an additional cause-level deduplication pass; cross-node duplicate screening has already been applied in Stage~2a. Individual cause states and current-episode frequencies remain unchanged.
\end{quote}

\end{document}